\documentclass[11pt, letterpaper, logo, onecolumn, copyright]{main}

\usepackage[authoryear, sort&compress, round]{natbib}

\usepackage[inkscapeformat=png]{svg}

\usepackage[most, breakable, skins]{tcolorbox}

\tcbuselibrary{skins}
\usepackage{lipsum}
\usepackage{tabularx}
\usepackage{afterpage}
\usepackage{booktabs}
\usepackage{subcaption}
\usepackage{makecell}
\usepackage{multirow}
\usepackage{bm}
\usepackage{multicol}
\usepackage{array}
\usepackage{float}
\usepackage{listings, listings-rust}
\usepackage{fontawesome5}
\usepackage{hyperref}
\usepackage{amssymb,graphicx}
\usepackage[dvipsnames]{xcolor}
\usepackage{cleveref}
\usepackage{longtable}
\usepackage{pdflscape}
\usepackage{adjustbox}
\usepackage{nicematrix}
\usepackage{CJKutf8}
\usepackage{ragged2e}
\usepackage{colortbl}
\usepackage{enumitem}
\usepackage[ruled,linesnumbered]{algorithm2e}
\usepackage{pifont}
\usepackage[htt]{hyphenat}
\usepackage{amsmath}
\usepackage{amsthm}  
\usepackage{mathrsfs} 
\usepackage{circledsteps}
\usepackage{diagbox}
\usepackage{bookmark}

\usepackage{wrapfig}
\usepackage{xspace}
\usepackage{tikz}
\usepackage[normalem]{ulem}

\usepackage{pgfplots}
\usepackage{pgfplotstable}
\pgfplotsset{compat=1.18}

\definecolor{medgray55}{gray}{0.55}
\definecolor{medgray}{gray}{0.7}
\definecolor{litegray}{gray}{0.9}
\definecolor{gblue}{RGB}{210, 227, 252}
\definecolor{gred}{RGB}{250, 210, 207}
\definecolor{gyellow}{RGB}{254, 239, 195}
\definecolor{ggreen}{RGB}{206, 234, 214}
\definecolor{gorange}{RGB}{254, 223, 200}

\definecolor{gblue9}{RGB}{23, 78, 166}
\definecolor{gred9}{RGB}{165, 14, 14}
\definecolor{gyellow9}{RGB}{227, 116, 0}
\definecolor{ggreen9}{RGB}{13, 101, 45}
\definecolor{gorange9}{RGB}{176, 96, 0}

\definecolor{myblue}{rgb}{0,0,1}
\definecolor{myred}{rgb}{1,0,0}
\definecolor{mylightgray}{gray}{0.95}
\definecolor{myCite}{HTML}{1C4587}

\definecolor{highlightblue}{HTML}{185ABC}
\definecolor{cellHighlight}{HTML}{dbefff}

\usepackage{minitoc}

\noptcrule


\newcolumntype{L}[1]{>{\raggedright\let\newline\\\arraybackslash\hspace{0pt}}m{#1}}
\newcolumntype{C}[1]{>{\centering}m{#1}}

\newcolumntype{R}[1]{>{\raggedleft\let\newline\\\arraybackslash\hspace{0pt}}m{#1}}

\definecolor{ao}{rgb}{0.0, 0.0, 1.0}

\newcommand\vcent[1]{\vcenter{\hbox{#1}}}

\newcommand\loudspeaker[1][3]{\ensuremath{\vcent{\rule{.6ex}{.6ex}}\kern-.5ex
  \vcent{\scalebox{.6}[1]{\rotatebox[origin=center]{90}{$\blacktriangle$}}}
  \ifnum#1>0\relax\kern.05ex\vcent{\scalebox{.4}{\ttfamily)}}
  \ifnum#1>1\relax\kern-.4ex\vcent{\scalebox{.56}{\ttfamily)}}
  \ifnum#1>2\relax\kern-.55ex\vcent{\scalebox{.7}{\ttfamily)}}
  \fi\fi\fi}
}

\newcommand{\ie}{\textit{i.e.,} }
\newcommand{\eg}{\textit{e.g.,} }

\newcommand{\para}[1]{
    \medskip\noindent\textbf{#1}
}

\makeatletter
\renewcommand\subparagraph{
 \@startsection {subparagraph}{5}{\z@ }{3.25ex \@plus 1ex
 \@minus .2ex}{-1em}{\normalfont \normalsize \bfseries }}
\makeatother

\let\cite\citep
\hypersetup{
  citecolor = myCite,  
  linkcolor = myCite,   
  urlcolor  = myCite
}

\title{Enhancing LLM-based Search Agents via Contribution Weighted Group Relative Policy Optimization}
\author{
    \textbf{Junzhe Wang$^1$$^{* \dag}$, Zhiheng Xi$^1$$^*$, Yajie Yang$^1$,}  \\
    \textbf{Hao Luo$^1$, Shihan Dou$^1$, Tao Gui$^1$, Qi Zhang$^{1,2,3}$$^\dag$}
\\
$^1$Fudan University $^2$Shanghai Artificial Intelligence Laboratory \\ $^3$Shanghai Key Laboratory of Intelligent Information Processing \\
\texttt{jzwang24@m.fudan.edu.cn, qz@fudan.edu.cn} 
}
\vspace{-50pt}
\begin{abstract}
Search agents extend Large Language Models (LLMs) beyond static parametric knowledge by enabling access to up-to-date and long-tail information unavailable during pretraining.
While reinforcement learning has been widely adopted for training such agents, existing approaches face key limitations: process supervision often suffers from unstable value estimation, whereas outcome supervision struggles with credit assignment due to sparse, trajectory-level rewards.
To bridge this gap, we propose Contribution-Weighted GRPO (CW-GRPO), a framework that integrates process supervision into group relative policy optimization.
Instead of directly optimizing process rewards, CW-GRPO employs an LLM judge to assess the retrieval utility and reasoning correctness at each search round, producing per-round contribution scores. These scores are used to rescale outcome-based advantages along the trajectory, enabling fine-grained credit assignment without sacrificing optimization stability.
Experiments on multiple knowledge-intensive benchmarks show that CW-GRPO outperforms standard GRPO by 5.0\% on Qwen3-8B and 6.3\% on Qwen3-1.7B, leading to more effective search behaviors.
Additional analysis reveals that successful trajectories exhibit concentrated contributions in specific rounds, providing empirical insight into search agent tasks.
\end{abstract}

\begin{document}

\doparttoc
\faketableofcontents

\begingroup
  \renewcommand\thefootnote{}
  \footnote{\textsuperscript{*}Equal contribution.
            \textsuperscript{\dag}Corresponding authors.}
    \footnote{\textsuperscript{1}Our code is available at \url{https://github.com/zsxmwjz/CW-GRPO}.}
  \addtocounter{footnote}{-1}
\endgroup

\vspace{-30pt}
\maketitle

\begin{figure}[!ht]
\begin{center}
\vspace{-10pt}
\includegraphics[width=0.6\linewidth]{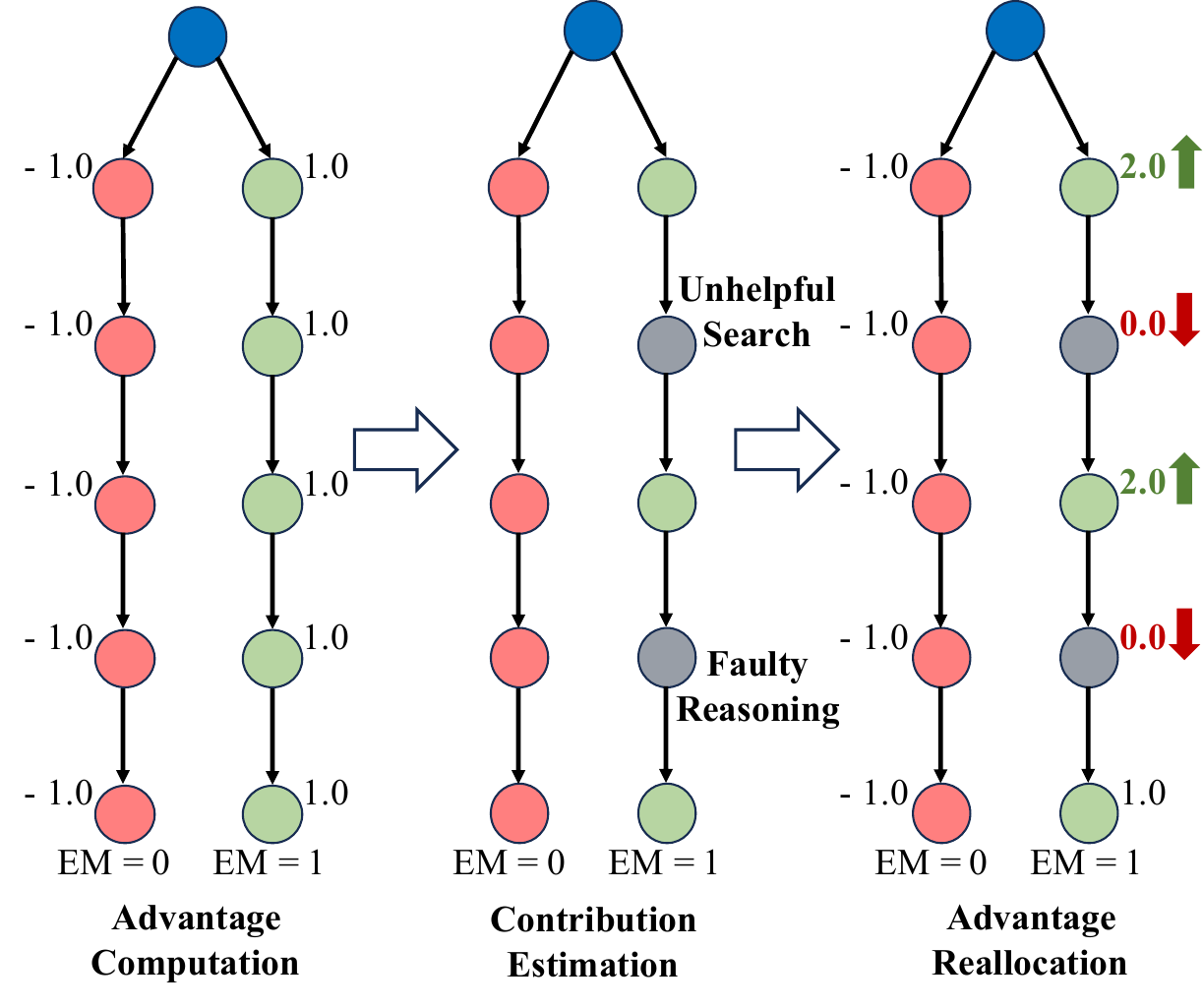}
\end{center}
\vspace{-16pt}
\caption{CW-GRPO assesses the quality of each search round within successful trajectories and to concentrate the learning advantage on high-quality search rounds, thereby enabling round-level credit assignment.}
\label{fig:overview}
\end{figure}

\section{Introduction}
\label{sec:introduction}

Search agents empower large language models (LLMs) to move beyond fixed parametric knowledge by providing access to real-time information and specialized facts absent from pre-training \citep{li2025searcho1agenticsearchenhancedlarge,zhang2025deep}.
Through iterative retrieval of external evidence and integration into the reasoning process, search agents ground model predictions in verifiable sources and substantially enhance the factual reliability of LLM-based systems  \citep{zhao2026retrieval,li2025matching,zhou2024trustworthiness,wang2025actingreasoningmoreteaching}.
Consequently, they have become a key role in enabling LLMs to perform knowledge-intensive and fact-sensitive tasks \citep{glass2022re2g,sawarkar2024blended}.

While reinforcement learning (RL) offers a significant paradigm for training such agents, current RL approaches suffer from fundamental limitations \citep{jin2025empiricalstudyreinforcementlearning}.
Depending on the level of supervision, they can be broadly categorized into process supervision and outcome supervision \citep{uesato2022solving}. 
Process supervision \citep{lightman2023letsverifystepstep,zhu2025chain,wang2025stepsearchignitingllmssearch} assigns round-level rewards and is commonly optimized with actor–critic methods such as Proximal Policy Optimization \citep{schulman2017proximalpolicyoptimizationalgorithms}, but learning reliable critics over diverse intermediate states is unstable and frequently leads to biased advantages and brittle training \citep{liu2024improvingmultistepreasoningabilities,kazemnejad2025vinepporefiningcreditassignment}.
Outcome supervision, by contrast, relies solely on final answer correctness, which makes it difficult to attribute task success to individual search rounds, leading to a \textbf{credit assignment problem} and obscuring the uneven contributions of intermediate decisions.
Recent advances, such as Group Relative Policy Optimization \citep{shao2024deepseekmathpushinglimitsmathematical}, improve training stability and memory efficiency, making outcome-supervised optimization more practical; however, they \citep{jin2025searchr1trainingllmsreason,wang2025actingreasoningmoreteaching} do not alter the sparsity of the reward signal, and the credit assignment issue remains unresolved.

To address the aforementioned challenges, we propose \textbf{C}ontribution-\textbf{W}eighted \textbf{GRPO (CW-GRPO)}, which reformulates process supervision as a problem of modulating outcome-derived advantages, rather than directly optimizing process rewards.
Specifically, CW-GRPO retains the stability of the standard GRPO by computing outcome advantages through group-relative comparisons.
We use an LLM judge to assess each search round based on retrieval utility and reasoning correctness.
These metrics are synthesized into a contribution score, based on which serves as a scaling factor to redistribute the outcome advantage across the trajectory.
By avoiding direct comparison of process rewards, CW-GRPO incorporates process supervision into the GRPO framework, amplifying learning signals for informative rounds while suppressing those for redundant or unproductive ones, as illustrated in Figure~\ref{fig:overview}.

Empirical results across multiple knowledge-intensive benchmarks demonstrate that CW-GRPO consistently outperforms both outcome-supervised and process-supervised baselines, enabling more effective search behaviors.
Specifically, CW-GRPO achieves significant performance gains over standard GRPO, with relative improvements of $5.0\%$ on Qwen3-8B and $6.3\%$ on Qwen3-1.7B.
The contributions of this work are summarized as follows:

\begin{itemize}[leftmargin=*]
    \item \textbf{A Reframing of Process Supervision}: We introduce a new perspective that treats process supervision for search agents as advantage reallocation guided by round contribution, rather than explicit process reward estimation.
    \item \textbf{Contribution-Weighted GRPO}: We propose a new optimization framework that brings process-level credit assignment into GRPO without requiring unstable value functions.
    \item \textbf{Empirical Characterization of Search Contribution}: We provide empirical evidence showing a structural characteristic in search agent tasks: the contribution to task success is highly concentrated in informative rounds rather than being uniformly distributed.
\end{itemize}

\section{Related Work}
\para{Agentic Search.} 
Large language models encode knowledge in static parameters, limiting access to up-to-date and long-tail knowledge. Retrieval-augmented generation (RAG) addresses this by enabling models to access external knowledge sources 
\citep{zhang2026opennoveltyllmpoweredagenticverifiable,zhao2026retrieval,li2025matching,zhou2024trustworthiness}. 
Pioneering retrieval-augmented reasoning works such as IRCoT \citep{trivedi-etal-2023-interleaving} and FLARE \citep{jiang-etal-2023-active} use prompts to guide iterative reasoning and reduce model uncertainty, and Self-RAG \citep{asai2024selfrag} adopts supervised fine-tuning with reflection tokens for adaptive retrieval. More recently, agentic search enables LLMs to interact with real-world search engines for dynamic information access.
WebGPT \citep{nakano2022webgptbrowserassistedquestionansweringhuman} pioneers this direction using RLHF. Search-o1 \citep{li2025searcho1agenticsearchenhancedlarge} integrates agentic search into large reasoning models via prompting, while Search-R1 \citep{jin2025searchr1trainingllmsreason}, R1-Searcher \citep{song2025r1searcherincentivizingsearchcapability} and MM-Doc-R1 \citep{lin2026mmdocr1trainingagentslong} adopt RL for end-to-end training with outcome-based rewards. However, outcome supervision assigns uniform credit across search rounds, failing to distinguish pivotal searches from redundant ones.

\para{Outcome Supervision and Process Supervision.} A key challenge in training search agents lies in credit assignment across multi-round search trajectories.
\emph{Outcome supervision}, widely adopted in RL methods like PPO \citep{schulman2017proximalpolicyoptimizationalgorithms} and GRPO \citep{shao2024deepseekmathpushinglimitsmathematical}, assigns rewards only based on the final answer, treating all immediate rounds equally and ignoring their unequal contributions.
\emph{Process supervision} addresses this via round-level feedback for finer-grained credit assignment.
Early approaches to process supervision rely on learned Process Reward Models (PRMs), but often suffer from costly round-level annotation requirements and limited out-of-domain generalizability \citep{lightman2023letsverifystepstep,huang-etal-2025-empirical}.
To avoid training explicit reward models, recent works increasingly leverage frozen LLMs as external evaluators to assess intermediate rounds. These methods use LLM-based judges to assess reasoning quality and then assign round-level rewards, enabling process supervision without training dedicated PRMs \citep{zhang2025criticsearchfinegrainedcreditassignment}.
These approaches typically optimize judge-produced signals directly as rewards, making them sensitive to evaluation noise and calibration errors. In contrast, CW-GRPO uses LLM-derived signals only to redistribute outcome-based advantages, avoiding direct optimization over noisy intermediate rewards and thereby enabling more stable training.

\section{Contribution-Weighted GRPO}

\begin{figure}[t]
\begin{center}
\vspace{-10pt}
\includegraphics[width=0.98\linewidth]{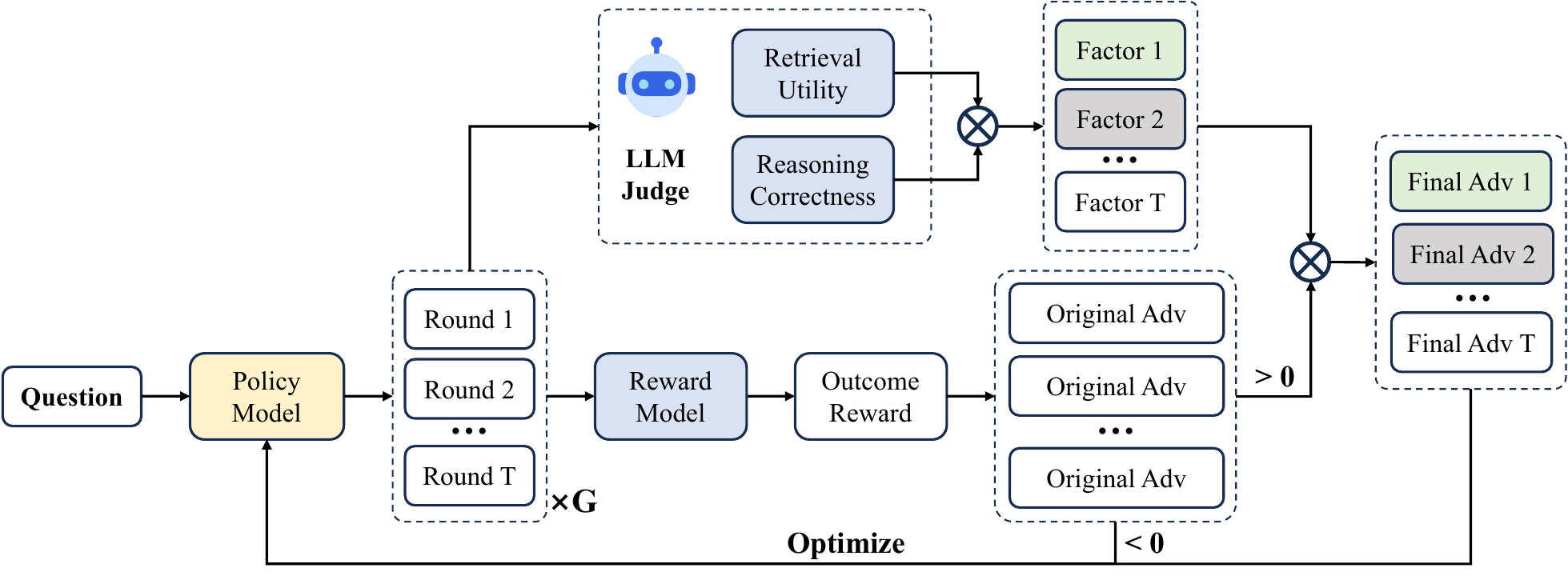}
\end{center}
\vspace{-16pt}
\caption{CW-GRPO employs an LLM judge to assess both the retrieval utility and the reasoning correctness at each individual search round. Based on these two signals, it computes a contribution factor for every search round, which is subsequently applied to the original advantage obtained from group-relative comparisons, thereby producing the final round advantage. As a result, training signals from highly contributive rounds are amplified, while those from low-contribution rounds are attenuated.}
\label{fig:method}
\end{figure}

As motivated in Section~\ref{sec:introduction}, current search agents suffer from the credit assignment problem where sparse outcome rewards cannot distinguish between pivotal and redundant search rounds.
To address this, we propose Contribution-Weighted Group Relative Policy Optimization (CW-GRPO).
Unlike traditional process supervision that requires learning a fallible critic, CW-GRPO reframes process-level signals as dynamic scaling factors that redistribute trajectory-level advantages.
The overall framework and its core mechanism of round-level credit assignment are presented in Figure~\ref{fig:method}.

\subsection{Task Formulation}

We consider a search agent policy $\pi_\theta$ tasked with answering a question q. The agent interacts with an environment (search engine) through a trajectory $\tau$ of $T$ rounds:
\begin{equation}
  \label{eq:trajectory}
  \small
  \tau = \big( (s^1, a^1), (s^2, a^2), \dots, (s^T, a^T) \big),
\end{equation}
where for each round $t < T$,  the action $a^t$ consists of a reasoning chain followed by an invocation of the search tool. The environment then returns retrieved documents to form the state $s^{t+1}$. The last action $a^T$ is the final answer.

Training proceeds in groups. For each question, we sample a group of $G$ trajectories $\{\tau_i\}_{i=1}^G$ under the current policy. Each trajectory $\tau_i$ receives a scalar outcome reward $R_i$ based on the exact match (EM) of its final answer. As in GRPO, these outcome rewards are only used for relative comparison within the group.

\subsection{Outcome-Level Advantage}

Following the GRPO framework, we first establish a baseline for performance by comparing trajectories within the same group. This avoids the need for a value-function critic. For each trajectory $\tau_i$, we compute a normalized outcome advantage:
\begin{equation}
  \label{eq:grpo_adv}
  \small
  A_i^O = \frac{R_i - \text{mean}\{R_i\}|_{i=1}^G} {\text{std}\{R_i\}|_{i=1}^G}.
\end{equation}
While $A_i^O$ effectively identifies which trajectories are better than others, it remains temporally coarse, assigning the same credit to every search round within a successful trajectory, regardless of whether a specific round was informative or distracting.

\subsection{Round-Level Contribution Estimation}

The effectiveness of CW-GRPO fundamentally depends on accurately identifying pivotal rounds within a search trajectory. Accordingly, we decompose the quality of each round into two orthogonal binary signals: \textbf{retrieval utility} and \textbf{reasoning correctness}.
Crucially, we adopt a \emph{conjunctive} formulation, under which a round is considered contributive only if it satisfies both criteria, namely retrieving novel, task-relevant evidence and maintaining logically consistent reasoning.
This design reflects an inherent structural property of multi-round search: effective task solving requires not only the acquisition of useful information but also its correct interpretation in context. Retrieval without sound reasoning may lead to incorrect interpretation or misuse of useful evidence; conversely, correct reasoning without informative retrieval fails to make meaningful progress toward solving the task.
The conjunctive gating mechanism thus serves as a conservative filter, isolating rounds whose contributions are causally aligned with task success, thereby providing a low-noise and more reliable signal for advantage reallocation.

\subsubsection{Signal Definition and Rubrics}

To ensure the robustness of the LLM-derived signals, we employ a \textbf{rubric-based assessment} where an LLM judge is guided by a set of predefined criteria.
For each search round $t < T_i$ in trajectory $\tau_i$, the values of the two signals are determined as follows:

\begin{itemize}[leftmargin=*]
    \item \textbf{Retrieval utility} $u_i^t \in \{0,1\}$: This signal assesses the information gained from external sources.
    A round is assigned $u_i^t=1$ only if the retrieved documents contain novel, task-relevant evidence that was not present in the context of previous rounds.
    This prevents the agent from being rewarded for redundant or circular retrieval queries.
    \item \textbf{Reasoning correctness} $v_i^t \in \{0,1\}$: This signal estimates the internal consistency of the agent.
    A round is assigned $v_i^t=1$ if the reasoning chain correctly interprets the current context and maintains a logical path toward the final answer.
    This ensures that the agent is not getting the right answer for the wrong reasons.
\end{itemize}

The conjunctive contribution signal is defined as the logical product $p_i^t = u_i^t \cdot v_i^t$.
By enforcing a binary bottleneck, we apply a discrete gate to each search round: it either contributes a necessary building block to the solution or it does not.

\subsubsection{Judge Calibration and Reliability}

To ensure that the LLM judge aligns with expert intuition, we conducted a rigorous calibration process.
We manually annotated a subset of trajectories (comprising 97 distinct search rounds) to serve as a gold standard.

The rubrics were iteratively refined to minimize ambiguity.
In our final setup, the LLM judge achieved a \textbf{95\% consensus rate with human expert annotations} on both retrieval utility and reasoning correctness.
This high level of agreement provides a reliable foundation for the subsequent advantage redistribution, ensuring that policy gradients are directed by high-fidelity process signals.
Additional calibration details are deferred to Appendix~\ref{cali_details}.

\subsection{Adaptive Weighting Mechanism}
\label{sec:weighting_mechanism}

We then convert these raw signals into a normalized \textbf{contribution weight} $c_i^t$. Importantly, we treat successful and failed trajectories differently, as a robustness-oriented design.

For \textbf{successful trajectories} ($R_i = 1$), rounds that introduce new task-relevant evidence and maintain logically consistent reasoning are typically associated with the final successful outcome. A temperature-controlled softmax is therefore adopted to emphasize high-contribution rounds:
\begin{equation}
        \label{eq:cont_success}
        \small
        c_i^t = \frac{\exp\big(\alpha p_i^t\big)} {\sum_{t'=1}^{T_i-1} \exp\big(\alpha p_i^{t'}\big)}, \quad t < T_i, \text{ if } R_i = 1,
\end{equation}
where $\alpha$ is a hyperparameter controlling the "sharpness" of the redistribution. A higher $\alpha$ forces the model to learn primarily from the most pivotal rounds.

For \textbf{failed trajectories} ($R_i = 0$), round-level attribution is substantially more ambiguous. Many intermediate rounds exhibit reasonable behavior, such as attempting to retrieve missing evidence or issuing follow-up queries for verification. However, these rounds may still fail to obtain useful information because the required knowledge is not covered by the corpus or cannot be retrieved due to limitations of the retriever.
In such cases, the failure cannot be readily attributed to identifiable errors in the agent’s decisions at specific rounds. As a result, assigning differentiated weights based on round-level signals is unreliable and can introduce spurious supervision. A more detailed analysis of such failure patterns is provided in Appendix~\ref{sec:failure_modes}.
To mitigate this issue, we assign a uniform contribution across all rounds:
\begin{equation}
        \label{eq:cont_fail}
        \small
        c_i^t = \frac{1}{T_i - 1}, \quad t < T_i, \text{ if } R_i = 0.
\end{equation}
This design preserves the stability of outcome-based learning while avoiding noisy round-level credit assignment under ambiguous attribution. Failed trajectories are still utilized through outcome-level comparison, and credit redistribution is applied only when attribution is reliable.

\noindent By construction, $\sum_{t=1}^{T_i-1} c_i^t = 1$.

\subsection{Advantage Reallocation and Optimization}

The core of CW-GRPO is the reallocated advantage $A_i^t$. For all search rounds $t < T_i$ of trajectory $\tau_i$, we scale the outcome advantage by the normalized contribution. And for the final answer round $t = T_i$, we directly use the outcome advantage:
\begin{equation}
  \label{eq:cw-grpo_adv}
  \small
  A_i^t = A_i^O \cdot c_i^t \cdot (T_i - 1), \text{ if } t < T_i;\quad A_i^{T_i} = A_i^O.
\end{equation}
The factor $(T_i - 1)$ keeps the total learning signal magnitude of the trajectory unchanged, \ie
\begin{equation}
  \label{eq:avg_adv_cw-grpo}
  \small
  \frac 1 {(T_i-1)}\sum_{t=1}^{T_i-1} A_i^t = A_i^O.
\end{equation}

\noindent Finally, we optimize the policy using the clipped surrogate objective:
\begin{equation}
  \label{eq:loss}
  \small
  \mathcal{L}(\theta) =
- \mathbb{E}\Big[
\min\big(
r A,
\text{clip}(r, 1-\epsilon, 1+\epsilon) A
\big)\Big],
\end{equation}
where $\epsilon$ is a hyper-parameter for clipping, $r=\frac {\pi_\theta(a \mid s)} {\pi_{\theta_{old}}(a \mid s)} = \exp(\log \pi_\theta - \log \pi_{\theta_{old}})$ is the importance sampling ratio.

In summary, CW-GRPO seamlessly integrates process supervision into the GRPO framework and achieves effective credit assignment across iterative search rounds by reallocating trajectory-level advantages based on each round’s contribution, rather than estimating absolute intermediate rewards.
This design preserves the training stability of group-relative optimization while ensuring the policy learns more effectively from the specific decisions that drive task success.

\section{Setup}

\subsection{Datasets and Evaluation Metrics}

We evaluate our method on two categories of knowledge-intensive tasks:
(1) \textbf{General Question Answering}: Natural Questions (NQ) \citep{kwiatkowski-etal-2019-natural}, TriviaQA \citep{joshi2017triviaqalargescaledistantly}, and PopQA \citep{mallen2023trustlanguagemodelsinvestigating}.
(2) \textbf{Multi-Hop Question Answering}: HotpotQA \citep{yang2018hotpotqadatasetdiverseexplainable}, 2WikiMultiHopQA \citep{ho2020constructingmultihopqadataset}, Musique \citep{trivedi2022musiquemultihopquestionssinglehop}, and Bamboogle \citep{press2023measuring}.
We merge the training sets of NQ and HotpotQA as the training data.
To mitigate the impact of decoding variance, we use Avg@4 Exact Match (EM) as the evaluation metric, averaging the EM scores over four sampled responses per task.

\para{Hard-Case Evaluation Set.} In standard settings, models often bypass external search by leveraging internal parametric knowledge, which obscures their true agentic search proficiency.
In light of this, we adopt a more challenging protocol using \textbf{AgentGym-SearchQA-test}, a 400-sample test set from AgentGym-RL \citep{xi2025agentgymrltrainingllmagents}. These samples are specifically filtered from cases where the Qwen2.5-72B-Instruct \citep{qwen2.5} model failed, representing the "hardest" distribution for current LLMs.
This setup shifts the focus from simple parametric recall to complex information-seeking and reasoning capabilities, explicitly emphasizing whether agents retrieve and utilize external evidence rather than relying on parametric knowledge.

\para{Strict Retrieval Constraint.} To further emphasize the model's search agentic capabilities, we include a specific instruction in the system prompt: the model is prohibited from using any parametric knowledge beyond basic common sense.

\subsection{Baselines}

We compare our approach against two categories of search agents:
(1) \textbf{Outcome-Supervised Methods}: Search-R1-PPO and Search-R1-GRPO~\citep{jin2025searchr1trainingllmsreason}, which optimize the final answer via PPO and GRPO respectively.
(2) \textbf{Process-Supervised Methods}: R3-RAG \citep{li2025r3} and MT-PPO \citep{wei2025reinforcingmultiturnreasoningllm}, which utilize explicit process rewards and PPO frameworks.

We also include results from state-of-the-art closed-source models (GPT-5.1 \citep{singh2025openaigpt5card}, Gemini-3-Pro-Preview \citep{google2025gemini3pro}, Qwen3-Max \citep{qwen3max}) as strong upper-bound references, alongside powerful open-source models (DeepSeek-V3.2 \citep{deepseekai2025deepseekv32pushingfrontieropen}, Qwen3-32B \citep{qwen3technicalreport}) for comparison.

\para{Note on Comparison:} Because of the uniquely challenging test set and the strict use of a no-parametric-knowledge prompt, the baseline performance reported in this paper differs from that reported in the original publications.

\subsection{Implementation Details}

We use Qwen3-8B and Qwen3-1.7B \citep{qwen3technicalreport} as base models due to their tool-calling capabilities.
Training is conducted using the veRL \citep{sheng2024hybridflow} framework.
The maximum context length is set to 9192 tokens, with up to 10 interaction rounds.
We employ SGLang \citep{zheng2024sglang} as the inference engine for rollout, setting both temperature and top\_p to 1.0.

\begin{table}[b]
    \centering
    \resizebox{0.75\textwidth}{!}{
    \begin{tabular}{l|ccc}
      \toprule
      \toprule
      \textbf{Parameter} 
& \textbf{\makecell{GRPO / CW-GRPO}} 
& \textbf{\makecell{PPO-based (Actor)}} 
& \textbf{\makecell{PPO-based (Critic)}} \\
      \midrule
      Learning Rate & $1\times10^{-6}$ & $1\times10^{-6}$ & $1\times10^{-5}$ \\
      Warmup Ratio & 0.285 & 0.285 & 0.015 \\
      Training Steps & 200 & 200 & 200 \\
      Batch Size & 32 & 128 & 128 \\
      Group Size & 4 & 1 & 1 \\
      GAE $\gamma / \lambda$ & - & 1.0/1.0 & 1.0/1.0 \\
      \bottomrule
      \bottomrule
    \end{tabular}
    }
    \caption{Detailed training hyperparameters of CW-GRPO and baselines.}
    \label{tab:hyper}
\end{table}

\para{Retrieval Infrastructure.} We use the wiki-18-corpus dataset released by Search-R1 \citep{jin2025searchr1trainingllmsreason}, derived from the 2018 Wikipedia dump, as the knowledge source, and E5-base-v2 \citep{wang2022text} as the retriever, with the number of retrieved documents $k$ set to 3.

\para{Training Hyperparameters.} For all RL-based experiments, we use a KL-divergence regularization coefficient $\beta = 0.001$ and a clipping ratio $\epsilon = 0.2$.
Specifically, for our proposed CW-GRPO, we apply GPT-oss-120B \citep{openai2025gptoss120bgptoss20bmodel} as the judge model and set $\alpha = \infty$. This configuration ensures that for trajectories resulting in task success, the learning signals during search iterations are exclusively concentrated on those rounds where the conjunctive contribution signal is $1$.
Other specific parameters are detailed in Table~\ref{tab:hyper}.
The hyperparameter differences between GRPO-based and PPO-based methods stem from their distinct algorithmic characteristics (intra-group relative comparisons vs. independent trajectory optimization), and we ensure fairness by normalizing the total number of sampled trajectories per training step.

\section{Experimental Results}

\begin{table*}[t]
    \centering
    \resizebox{\textwidth}{!}{
    \begin{tabular}{l|ccc|cccc|c}
      \toprule
      \toprule
        \multicolumn{1}{c}{\multirow{2}{*}{\textbf{Method}}} & \multicolumn{3}{|c}{\textbf{General QA}} & \multicolumn{4}{|c}{\textbf{Multi-Hop QA}} & \multicolumn{1}{|c}{\multirow{2}{*}{\textbf{Overall}}} \\
        \cmidrule(lr){2-4}\cmidrule(lr){5-8}
          & NQ\textsuperscript{\textdagger} & TriviaQA\textsuperscript{*} & PopQA\textsuperscript{*} & HotpotQA\textsuperscript{\textdagger} & 2wiki\textsuperscript{*} & Musique\textsuperscript{*} & Bamboogle\textsuperscript{*} \\
            \midrule
            \rowcolor[rgb]{ 0.850, 0.890, 0.956} \multicolumn{9}{l}{\textbf{Proprietary Models}} \\
            GPT-5.1 & 31.0 & 60.0 & 38.5 & 36.0 & 43.5 & 13.8 & 39.0 & 34.44 \\
            Gemini-3-Pro-Preview & 31.0 & 66.0 & 32.5 & 42.0 & 38.5 & 14.0 & 42.0 & 35.00 \\
            Qwen3-Max & 31.5 & 62.5 & 30.5 & 39.5 & 49.0 & 18.5 & 42.5 & 36.56 \\
            \midrule
            \rowcolor[rgb]{ 0.850, 0.890, 0.956} \multicolumn{9}{l}{\textbf{Open-sourced Models}} \\
            DeepSeek-V3.2-Thinking & 24.5 & 47.0 & 22.0 & 30.0 & 24.0 & 12.3 & 29.5 & 25.19 \\
            Qwen3-32B & 20.5 & 53.0 & 22.0 & 28.0 & 33.5 & 8.5 & 30.0 & 25.50 \\
            \midrule
            \textbf{Qwen3-1.7B} & 13.5 & 34.5 & 18.0 & 11.0 & 6.0 & 2.5 & 4.5 & 11.56 \\
            Search-R1-PPO & 19.5 & \textbf{50.5} & 25.0 & 25.5 & 24.0 & 5.0 & 14.0 & 21.06 \\
            Search-R1-GRPO & 18.0 & 45.5 & 25.0 & \textbf{27.0} & 22.5 & 6.8 & 16.5 & 21.00 \\
            R3-RAG & 20.0 & 47.5 & \textbf{25.5} & 22.0 & 18.5 & 5.3 & 13.5 & 19.69 \\
            MT-PPO & \textbf{22.5} & 49.0 & 25.0 & 24.0 & \textbf{31.5} & 6.0 & 13.5 & 22.19 \\
            \textbf{CW-GRPO} & 22.0 & 47.5 & 25.0 & 24.0 & 28.5 & \textbf{7.0} & \textbf{17.5} & \textbf{22.31} \\
            \midrule
            \textbf{Qwen3-8B} & 23.5 & 53.5 & 22.5 & 27.5 & 27.0 & 10.0 & 30.0 & 25.50 \\
            Search-R1-PPO & 29.5 & 57.0 & 27.5 & \textbf{33.5} & 36.0 & 12.3 & 33.5 & 30.19 \\
            Search-R1-GRPO & 33.5 & 57.0 & 26.0 & \textbf{33.5} & 32.5 & 12.5 & 31.5 & 29.88 \\
            R3-RAG & 26.5 & \textbf{57.5} & 26.0 & 32.0 & 32.0 & 11.3 & 33.5 & 28.75 \\
            MT-PPO & 25.5 & 56.5 & 26.0 & 32.0 & \textbf{37.0} & 12.5 & 31.5 & 29.19 \\
            \textbf{CW-GRPO} & \textbf{35.5} & \textbf{57.5} & \textbf{28.5} & \textbf{33.5} & 33.0 & \textbf{13.0} & \textbf{37.0} & \textbf{31.38} \\
      \bottomrule
      \bottomrule
    \end{tabular}
    }
    \caption{Avg@4 Exact Match on seven subsets of AgentGym-SearchQA-test. \textdagger \ /\ * indicate in-domain and out-of-domain datasets. CW-GRPO achieves the best performance among methods built on the same backbone.}
    \label{tab:main}
\end{table*}

\subsection{Main Results}

Table \ref{tab:main} presents the main experimental results on hard knowledge-intensive QA benchmarks under no-parametric-knowledge constraints.

\para{Overall Performance.} Our experimental results demonstrate the robust superiority of CW-GRPO, which achieves state-of-the-art performance across both model scales, outperforming all outcome- and process-supervised baselines on the identical backbone.
On Qwen3-8B, CW-GRPO achieves an overall score of 31.38, representing a $5.0\%$ relative improvement over Search-R1-GRPO (29.88).
On Qwen3-1.7B, CW-GRPO improves the overall performance from 21.00 to 22.31, corresponding to a $6.3\%$ relative gain.
These results indicate that CW-GRPO scales favorably across model sizes and is particularly effective in low-capacity regimes, where efficient search credit assignment is critical.

\para{Comparison with Open- and Closed-Source Models.} With Qwen3-8B as the backbone, CW-GRPO surpasses all evaluated open-source models, including DeepSeek-V3.2-Thinking and Qwen3-32B, despite operating under a stricter evaluation setting.
However, a clear performance gap remains between CW-GRPO and large closed-source models such as GPT-5.1, Gemini-3-Pro-Preview, and Qwen3-Max, suggesting that stronger base models and proprietary training data still provide advantages under this challenging hard-case distribution.

\para{General QA vs. Multi-Hop QA.} CW-GRPO demonstrates consistent improvements on both General QA and Multi-Hop QA tasks compared to outcome-supervised and process-supervised baselines.
On Multi-Hop QA, CW-GRPO yields the most pronounced gains. On both Qwen3-1.7B and Qwen3-8B, it improves performance on 2WikiMultiHopQA, Musique, and Bamboogle simultaneously, indicating stronger long-horizon reasoning and evidence aggregation.
On General QA, CW-GRPO remains competitive across datasets. While it achieves the best overall General QA performance on Qwen3-8B, its improvement on Qwen3-1.7B is smaller than that of MT-PPO.
This suggests that for simpler single-hop questions, dense process supervision can still be effective for very small models, whereas CW-GRPO shows clearer advantages as reasoning depth increases.

\para{Summary.} Overall, these results demonstrate that CW-GRPO provides a robust and scalable improvement over existing outcome- and process-supervised search agents, particularly for multi-hop reasoning under strict retrieval constraints. The consistent gains across model sizes highlight the effectiveness of contribution-weighted optimization in guiding search behavior, even in extremely challenging evaluation settings.

\subsection{Sharpness of Advantage Reallocation}

Based on Qwen3-8B, we analyze the effect of the sharpness parameter $\alpha$, which controls how outcome advantages are redistributed across search rounds based on their estimated contributions.
Setting $\alpha=0$ yields uniform weighting over the search rounds and reduces the method to standard GRPO, whereas $\alpha=\infty$ corresponds to CW-GRPO, where the advantage is fully concentrated on rounds with both retrieval utility and correct reasoning.

As shown in Table~\ref{tab:ablation_alpha}, increasing $\alpha$ generally leads to improved overall performance, with the best overall result achieved at $\alpha=\infty$.
This trend suggests that, in successful search trajectories, the contribution to task success is highly uneven across rounds.
Rather than accumulating gradually, effective progress is driven by high-quality search rounds that exhibit retrieval utility and correct reasoning.
For search agent tasks, effective credit assignment requires acknowledging this highly concentrated contribution structure.

\subsection{Supervision Decomposition}

We ablate the two supervision signals in CW-GRPO by separately removing retrieval utility or reasoning correctness supervision while keeping all other components unchanged.

According to Table~\ref{tab:ablation_llmjudge}, removing either signal consistently degrades performance across both single-hop and multi-hop QA tasks, suggesting that neither signal alone is sufficient for reliable credit assignment during search.

Without retrieval utility supervision, rounds with coherent but uninformative reasoning can receive positive credit.
Without reasoning correctness supervision, rounds can be assigned positive contribution weights for retrieving relevant documents even when they are incorrectly interpreted, weakening the link between evidence acquisition and correct reasoning. 
In either setting, contribution weights may be assigned to rounds that do not make a substantive contribution to task completion, introducing noise into advantage reallocation and harming overall performance.

These results highlight the necessity of joint supervision: effective search requires both acquiring novel information and correctly reasoning over it.

\begin{table*}[t]
\small
    \centering
    \resizebox{\textwidth}{!}{
    \begin{tabular}{c|ccc|cccc|c}
      \toprule
      \toprule
        \multicolumn{1}{c}{\multirow{2}{*}{\quad$\alpha$\quad}} & \multicolumn{3}{|c}{\textbf{General QA}} & \multicolumn{4}{|c}{\textbf{Multi-Hop QA}} & \multicolumn{1}{|c}{\multirow{2}{*}{\textbf{Overall}}} \\
        \cmidrule(lr){2-4}\cmidrule(lr){5-8}
          & NQ & TriviaQA & PopQA & HotpotQA & 2wiki & Musique & Bamboogle \\
            \midrule
            0 & 33.5 & 57.0 & 26.0 & 33.5 & 32.5 & 12.5 & 31.5 & 29.88 \\
            1 & 27.0 & 58.0 & 28.0 & 32.5 & 35.0 & 12.8 & 31.5 & 29.69 \\
            3 & 33.0 & 58.5 & 31.0 & 29.5 & 36.0 & 13.0 & 31.0 & 30.63 \\
            5 & 27.0 & 58.5 & 27.0 & 35.0 & 39.0 & 12.0 & 33.5 & 30.50 \\
            $\infty$ & 35.5 & 57.5 & 28.5 & 33.5 & 33.0 & 13.0 & 37.0 & 31.38 \\
      \bottomrule
      \bottomrule
    \end{tabular}
    }
    \caption{Effect of the sharpness parameter $\alpha$ on search agent performance. Larger $\alpha$ value generally yields better overall accuracy, indicating that concentrating learning signals on high-contribution search rounds is beneficial.}
    \label{tab:ablation_alpha}
\end{table*}

\begin{table*}[t]
\small
    \centering
    \resizebox{\textwidth}{!}{
    \begin{tabular}{l|ccc|cccc|c}
      \toprule
      \toprule
        \multicolumn{1}{c}{\multirow{2}{*}{\textbf{Method}}} & \multicolumn{3}{|c}{\textbf{General QA}} & \multicolumn{4}{|c}{\textbf{Multi-Hop QA}} & \multicolumn{1}{|c}{\multirow{2}{*}{\textbf{Overall}}} \\
        \cmidrule(lr){2-4}\cmidrule(lr){5-8}
          & NQ & TriviaQA & PopQA & HotpotQA & 2wiki & Musique & Bamboogle \\
            \midrule
            CW-GRPO & 35.5 & 57.5 & 28.5 & 33.5 & 33.0 & 13.0 & 37.0 & 31.38 \\
            wo-retrieval & 27.0 & 53.5 & 26.0 & 31.0 & 37.5 & 12.0 & 34.5 & 29.19 \\
            wo-reasoning & 23.5 & 55.5 & 31.0 & 34.0 & 28.5 & 12.3 & 31.5 & 28.56 \\
      \bottomrule
      \bottomrule
    \end{tabular}
    }
    \caption{Ablation on contribution supervision in CW-GRPO. Removing either retrieval utility or reasoning correctness supervision degrades performance, showing both are necessary for effective credit assignment.}
    \label{tab:ablation_llmjudge}
\end{table*}

\subsection{Training Dynamics}

As shown in Figure~\ref{fig:training}, the optimization remains stable across training, with no signs of collapse.
The Exact Match score steadily improves, accompanied by consistent upward trends in both retrieval utility and reasoning correctness, indicating progressive improvement in the agent’s search behavior.
These results show that CW-GRPO achieves stable optimization while improving both final accuracy and the quality of the search process.

We further analyze the effect of the conjunctive gate mechanism during training.
Among all search rounds in successful trajectories, approximately 13.84\% (2.2k out of 15.8k) are assigned zero contribution, indicating that the gate meaningfully reshapes the distribution of learning signals rather than acting as a near-identity transformation.

\subsection{Case Study}

\begin{figure*}[t]
    \centering
    \includegraphics[width=0.98\textwidth]{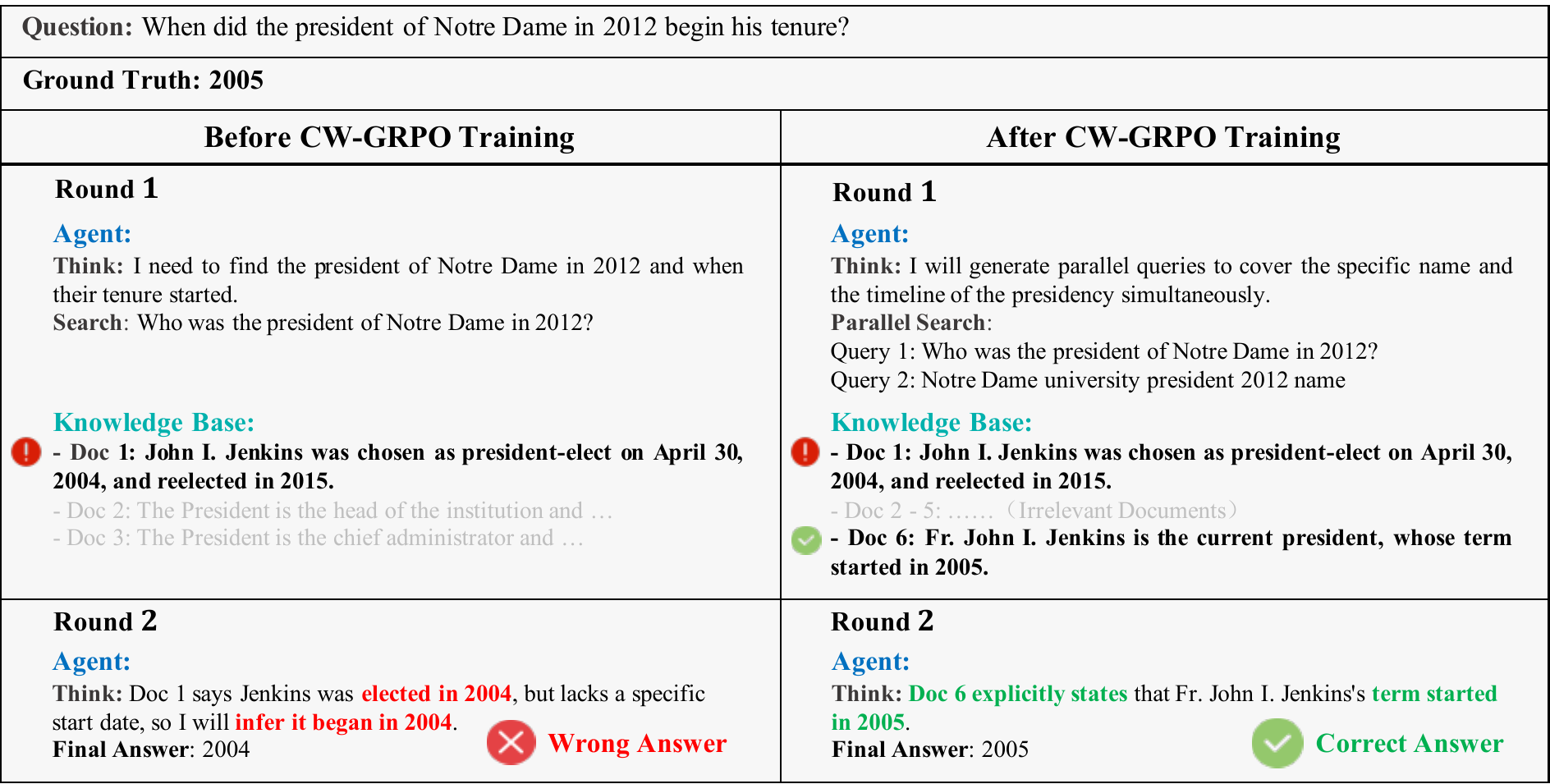}
    \caption{Case study of CW-GRPO on Qwen3-8B.}
    \label{fig:case_study}
\end{figure*}

Figure~\ref{fig:case_study} compares the search trajectories of Qwen3-8B before and after CW-GRPO training on a factual question regarding John I. Jenkins’s tenure start.
The untrained model issues only a single query and fails to distinguish between the election year and the actual commencement of his term, leading to an incorrect answer based on this temporal confusion.

In contrast, the CW-GRPO-trained model adopts a qualitatively different strategy by issuing multiple parallel queries in the first round.
This expanded search coverage allows the agent to recall a broader set of documents, including explicit evidence that Jenkins’s term began in 2005.
This case illustrates that CW-GRPO not only improves retrieval utility through parallelization but also strengthens the agent’s resistance to distracting evidence. By learning to prioritize direct answer-bearing evidence over superficially relevant but misleading signals, the model achieves more robust and grounded search-based decision making.

\begin{figure*}[t]
    \centering
    \includegraphics[width=\textwidth]{figures/training_dynamics.png}
    \caption{Training curves of CW-GRPO on Qwen3-8B. We report the exponential moving averages of Exact Match, retrieval utility, and reasoning correctness over training steps.}
    \label{fig:training}
\end{figure*}

\section{Discussion: Generalization Across Tasks and Models}

CW-GRPO is motivated by the general principle of redistributing outcome-derived advantages according to round-level contribution estimates.
This principle is broadly applicable beyond the specific experimental settings considered in this work.

\para{Task generalization.} CW-GRPO can be applied to multi-round agentic tasks where the contributions of intermediate rounds can be reasonably estimated.
In long-horizon decision-making environments(\eg Webshop~\citep{yao2022webshop}), individual actions should correspond to incremental progress toward latent subgoals.
This enables a judge to assess state improvements at each round and facilitates round-level credit assignment.

The framework also extends to single-round but structurally decomposable reasoning tasks. In settings such as mathematical reasoning(\eg MATH~\citep{hendrycks2021measuringmathematicalproblemsolving}), model outputs typically consist of multi-step derivations. When a trajectory fails, the error can often be localized to specific steps; these steps can then be assigned higher responsibility for failure, enabling a CW-GRPO variant based on step-level attribution.

However, CW-GRPO is less suitable for tasks where process contribution is ill-defined or inherently unstable. This includes highly subjective generation tasks (\eg open-ended creative writing) and perceptual generation tasks (\eg image synthesis), where holistic judgments dominate and assigning reliable credit to intermediate steps via low-variance signals is challenging.

\para{Model generalization.} CW-GRPO is model-agnostic and does not rely on a specific backbone architecture. Nevertheless, stronger base models tend to be more sensitive to the accuracy of round-level supervision, placing higher demands on the quality of the judge or rubric design, which may limit the practical gains of the method.

\section{Conclusion}
In this work, we propose CW-GRPO, which reallocates outcome advantages according to round-level contribution rather than explicit process rewards to tackle the credit assignment challenge in training LLM-based search agents.
CW-GRPO retains GRPO’s stability while enabling fine-grained credit assignment along search rounds.
Our evaluations show that CW-GRPO consistently outperforms baselines, and task success is closely associated with the high-quality rounds, highlighting the importance of modeling contribution concentration in search trajectories.
Overall, CW-GRPO offers a simple and effective way to incorporate process-level insights into outcome-supervised training for scalable agentic systems.

\section*{Limitations}

Despite the advances demonstrated by CW-GRPO in alleviating credit assignment under outcome supervision, several limitations remain.

First, CW-GRPO reallocates advantages only for successful trajectories.
For failed trajectories, it does not perform fine-grained credit assignment, as attributing task failure to specific search rounds remains challenging.
Consequently, CW-GRPO lacks explicit modeling of which intermediate decisions contribute to failure, leaving failure cases underutilized for process-aware learning.

Second, CW-GRPO adopts a conjunctive gating mechanism for modeling round-level contributions, which improves robustness to judging noise but limits the expressiveness of contribution representation.
In addition, the framework operates at the round level and does not naturally extend to token-level supervision, making finer-grained credit assignment beyond rounds hard to incorporate.

Finally, CW-GRPO relies on an external LLM judge for contribution estimation. While we find the associated computational cost to be manageable in our setting (see Appendix~\ref{sec:computational_cost}), this design still introduces additional inference cost and dependency on judge quality. Exploring lighter-weight supervision signals (\eg heuristic judges) may further improve scalability, especially in more resource-constrained or latency-sensitive scenarios.

\bibliography{main}

\begin{thebibliography}{49}
\providecommand{\natexlab}[1]{#1}
\providecommand{\url}[1]{\texttt{#1}}
\expandafter\ifx\csname urlstyle\endcsname\relax
  \providecommand{\doi}[1]{doi: #1}\else
  \providecommand{\doi}{doi: \begingroup \urlstyle{rm}\Url}\fi

\bibitem[Agarwal et~al.(2025)Agarwal, Ahmad, Ai, Altman, Applebaum, Arbus, Arora, Bai, Baker, Bao, Barak, Bennett, Bertao, Brett, Brevdo, Brockman, Bubeck, Chang, Chen, Chen, Cheung, Clark, Cook, Dukhan, Dvorak, Fives, Fomenko, Garipov, Georgiev, Glaese, Gogineni, Goucher, Gross, Guzman, Hallman, Hehir, Heidecke, Helyar, Hu, Huet, Huh, Jain, Johnson, Koch, Kofman, Kundel, Kwon, Kyrylov, Le, Leclerc, Lennon, Lessans, Lezcano-Casado, Li, Li, Lin, Liss, Liu, Liu, Lu, Lu, Martinovic, McCallum, McGrath, McKinney, McLaughlin, Mei, Mostovoy, Mu, Myles, Neitz, Nichol, Pachocki, Paino, Palmie, Pantuliano, Parascandolo, Park, Pathak, Paz, Peran, Pimenov, Pokrass, Proehl, Qiu, Raila, Raso, Ren, Richardson, Robinson, Rotsted, Salman, Sanjeev, Schwarzer, Sculley, Sikchi, Simon, Singhal, Song, Stuckey, Sun, Tillet, Toizer, Tsimpourlas, Vyas, Wallace, Wang, Wang, Watkins, Weil, Wendling, Whinnery, Whitney, Wong, Yang, Yang, Yasunaga, Ying, Zaremba, Zhan, Zhang, Zhang, Zhang, and Zhao]{openai2025gptoss120bgptoss20bmodel}
OpenAI:~Sandhini Agarwal, Lama Ahmad, Jason Ai, Sam Altman, Andy Applebaum, Edwin Arbus, Rahul~K. Arora, Yu~Bai, Bowen Baker, Haiming Bao, Boaz Barak, Ally Bennett, Tyler Bertao, Nivedita Brett, Eugene Brevdo, Greg Brockman, Sebastien Bubeck, Che Chang, Kai Chen, Mark Chen, Enoch Cheung, Aidan Clark, Dan Cook, Marat Dukhan, Casey Dvorak, Kevin Fives, Vlad Fomenko, Timur Garipov, Kristian Georgiev, Mia Glaese, Tarun Gogineni, Adam Goucher, Lukas Gross, Katia~Gil Guzman, John Hallman, Jackie Hehir, Johannes Heidecke, Alec Helyar, Haitang Hu, Romain Huet, Jacob Huh, Saachi Jain, Zach Johnson, Chris Koch, Irina Kofman, Dominik Kundel, Jason Kwon, Volodymyr Kyrylov, Elaine~Ya Le, Guillaume Leclerc, James~Park Lennon, Scott Lessans, Mario Lezcano-Casado, Yuanzhi Li, Zhuohan Li, Ji~Lin, Jordan Liss, Lily Liu, Jiancheng Liu, Kevin Lu, Chris Lu, Zoran Martinovic, Lindsay McCallum, Josh McGrath, Scott McKinney, Aidan McLaughlin, Song Mei, Steve Mostovoy, Tong Mu, Gideon Myles, Alexander Neitz, Alex Nichol, Jakub
  Pachocki, Alex Paino, Dana Palmie, Ashley Pantuliano, Giambattista Parascandolo, Jongsoo Park, Leher Pathak, Carolina Paz, Ludovic Peran, Dmitry Pimenov, Michelle Pokrass, Elizabeth Proehl, Huida Qiu, Gaby Raila, Filippo Raso, Hongyu Ren, Kimmy Richardson, David Robinson, Bob Rotsted, Hadi Salman, Suvansh Sanjeev, Max Schwarzer, D.~Sculley, Harshit Sikchi, Kendal Simon, Karan Singhal, Yang Song, Dane Stuckey, Zhiqing Sun, Philippe Tillet, Sam Toizer, Foivos Tsimpourlas, Nikhil Vyas, Eric Wallace, Xin Wang, Miles Wang, Olivia Watkins, Kevin Weil, Amy Wendling, Kevin Whinnery, Cedric Whitney, Hannah Wong, Lin Yang, Yu~Yang, Michihiro Yasunaga, Kristen Ying, Wojciech Zaremba, Wenting Zhan, Cyril Zhang, Brian Zhang, Eddie Zhang, and Shengjia Zhao.
\newblock gpt-oss-120b \& gpt-oss-20b model card, 2025.
\newblock URL \url{https://arxiv.org/abs/2508.10925}.

\bibitem[Asai et~al.(2024)Asai, Wu, Wang, Sil, and Hajishirzi]{asai2024selfrag}
Akari Asai, Zeqiu Wu, Yizhong Wang, Avirup Sil, and Hannaneh Hajishirzi.
\newblock Self-{RAG}: Learning to retrieve, generate, and critique through self-reflection.
\newblock In \emph{The Twelfth International Conference on Learning Representations}, 2024.
\newblock URL \url{https://openreview.net/forum?id=hSyW5go0v8}.

\bibitem[DeepSeek-AI et~al.(2025)DeepSeek-AI, Liu, Mei, Lin, Xue, Wang, Xu, Wu, Zhang, Lin, Dong, Lu, Zhao, Deng, Xu, Ruan, Dai, Guo, Yang, Chen, Li, Zhou, Lin, Dai, Hao, Chen, Li, Zhang, Xu, Li, Liang, Wei, Zhang, Luo, Ji, Ding, Tang, Cao, Gao, Qu, Zeng, Huang, Li, Xu, Hu, Chen, Xiang, Yuan, Cheng, Zhu, Ran, Jiang, Qiu, Li, Song, Dong, Gao, Guan, Huang, Zhou, Huang, Yu, Wang, Zhang, Wang, Zhao, Yin, Guo, Luo, Ma, Wang, Zhang, Di, Xu, Zhang, Zhang, Tang, Zhou, Huang, Cong, Wang, Wang, Zhu, Li, Chen, Du, Xu, Ge, Zhang, Pan, Wang, Yin, Xu, Shen, Zhang, Liu, Lu, Zhou, Chen, Cai, Chen, Hu, Liu, Hu, Ma, Wang, Yu, Zhou, Pan, Zhou, Ni, Yun, Pei, Ye, Yue, Zeng, Liu, Liang, Pang, Luo, Gao, Zhang, Gao, Wang, Bi, Liu, Wang, Chen, Zhang, Nie, Cheng, Liu, Xie, Liu, Yu, Li, Yang, Li, Chen, Su, Pan, Lin, Fu, Wang, Zhang, Xu, Ma, Li, Li, Zhao, Sun, Wang, Qian, Yu, Zhang, Ding, Shi, Xiong, He, Zhou, Zhong, Piao, Wang, Chen, Tan, Wei, Ma, Liu, Yang, Guo, Wu, Wu, Cheng, Ou, Xu, Wang, Gong, Wu, Zou, Li, Xiong, Luo, You, Liu,
  Zhou, Wu, Ren, Zhao, Ren, Sha, Fu, Xu, Xie, Zhang, Hao, Gou, Ma, Yan, Shao, Huang, Wu, Li, Zhang, Xu, Wang, Gu, Zhu, Li, Zhang, Xie, Gao, Pan, Yao, Feng, Li, Cai, Ni, Xu, Li, Tian, Chen, Jin, Li, Zhou, Sun, Li, Jin, Shen, Chen, Song, Zhou, Zhu, Huang, Li, Zheng, Zhu, Ma, Huang, Xu, Zhang, Ji, Liang, Guo, Chen, Xia, Wang, Li, Zhang, Chen, Sun, Wu, Ye, Wang, Xiao, An, Wang, Sun, Wang, Tang, Zha, Zhang, Ju, Zhang, and Qu]{deepseekai2025deepseekv32pushingfrontieropen}
DeepSeek-AI, Aixin Liu, Aoxue Mei, Bangcai Lin, Bing Xue, Bingxuan Wang, Bingzheng Xu, Bochao Wu, Bowei Zhang, Chaofan Lin, Chen Dong, Chengda Lu, Chenggang Zhao, Chengqi Deng, Chenhao Xu, Chong Ruan, Damai Dai, Daya Guo, Dejian Yang, Deli Chen, Erhang Li, Fangqi Zhou, Fangyun Lin, Fucong Dai, Guangbo Hao, Guanting Chen, Guowei Li, H.~Zhang, Hanwei Xu, Hao Li, Haofen Liang, Haoran Wei, Haowei Zhang, Haowen Luo, Haozhe Ji, Honghui Ding, Hongxuan Tang, Huanqi Cao, Huazuo Gao, Hui Qu, Hui Zeng, Jialiang Huang, Jiashi Li, Jiaxin Xu, Jiewen Hu, Jingchang Chen, Jingting Xiang, Jingyang Yuan, Jingyuan Cheng, Jinhua Zhu, Jun Ran, Junguang Jiang, Junjie Qiu, Junlong Li, Junxiao Song, Kai Dong, Kaige Gao, Kang Guan, Kexin Huang, Kexing Zhou, Kezhao Huang, Kuai Yu, Lean Wang, Lecong Zhang, Lei Wang, Liang Zhao, Liangsheng Yin, Lihua Guo, Lingxiao Luo, Linwang Ma, Litong Wang, Liyue Zhang, M.~S. Di, M.~Y Xu, Mingchuan Zhang, Minghua Zhang, Minghui Tang, Mingxu Zhou, Panpan Huang, Peixin Cong, Peiyi Wang, Qiancheng Wang,
  Qihao Zhu, Qingyang Li, Qinyu Chen, Qiushi Du, Ruiling Xu, Ruiqi Ge, Ruisong Zhang, Ruizhe Pan, Runji Wang, Runqiu Yin, Runxin Xu, Ruomeng Shen, Ruoyu Zhang, S.~H. Liu, Shanghao Lu, Shangyan Zhou, Shanhuang Chen, Shaofei Cai, Shaoyuan Chen, Shengding Hu, Shengyu Liu, Shiqiang Hu, Shirong Ma, Shiyu Wang, Shuiping Yu, Shunfeng Zhou, Shuting Pan, Songyang Zhou, Tao Ni, Tao Yun, Tian Pei, Tian Ye, Tianyuan Yue, Wangding Zeng, Wen Liu, Wenfeng Liang, Wenjie Pang, Wenjing Luo, Wenjun Gao, Wentao Zhang, Xi~Gao, Xiangwen Wang, Xiao Bi, Xiaodong Liu, Xiaohan Wang, Xiaokang Chen, Xiaokang Zhang, Xiaotao Nie, Xin Cheng, Xin Liu, Xin Xie, Xingchao Liu, Xingkai Yu, Xingyou Li, Xinyu Yang, Xinyuan Li, Xu~Chen, Xuecheng Su, Xuehai Pan, Xuheng Lin, Xuwei Fu, Y.~Q. Wang, Yang Zhang, Yanhong Xu, Yanru Ma, Yao Li, Yao Li, Yao Zhao, Yaofeng Sun, Yaohui Wang, Yi~Qian, Yi~Yu, Yichao Zhang, Yifan Ding, Yifan Shi, Yiliang Xiong, Ying He, Ying Zhou, Yinmin Zhong, Yishi Piao, Yisong Wang, Yixiao Chen, Yixuan Tan, Yixuan Wei, Yiyang
  Ma, Yiyuan Liu, Yonglun Yang, Yongqiang Guo, Yongtong Wu, Yu~Wu, Yuan Cheng, Yuan Ou, Yuanfan Xu, Yuduan Wang, Yue Gong, Yuhan Wu, Yuheng Zou, Yukun Li, Yunfan Xiong, Yuxiang Luo, Yuxiang You, Yuxuan Liu, Yuyang Zhou, Z.~F. Wu, Z.~Z. Ren, Zehua Zhao, Zehui Ren, Zhangli Sha, Zhe Fu, Zhean Xu, Zhenda Xie, Zhengyan Zhang, Zhewen Hao, Zhibin Gou, Zhicheng Ma, Zhigang Yan, Zhihong Shao, Zhixian Huang, Zhiyu Wu, Zhuoshu Li, Zhuping Zhang, Zian Xu, Zihao Wang, Zihui Gu, Zijia Zhu, Zilin Li, Zipeng Zhang, Ziwei Xie, Ziyi Gao, Zizheng Pan, Zongqing Yao, Bei Feng, Hui Li, J.~L. Cai, Jiaqi Ni, Lei Xu, Meng Li, Ning Tian, R.~J. Chen, R.~L. Jin, S.~S. Li, Shuang Zhou, Tianyu Sun, X.~Q. Li, Xiangyue Jin, Xiaojin Shen, Xiaosha Chen, Xinnan Song, Xinyi Zhou, Y.~X. Zhu, Yanping Huang, Yaohui Li, Yi~Zheng, Yuchen Zhu, Yunxian Ma, Zhen Huang, Zhipeng Xu, Zhongyu Zhang, Dongjie Ji, Jian Liang, Jianzhong Guo, Jin Chen, Leyi Xia, Miaojun Wang, Mingming Li, Peng Zhang, Ruyi Chen, Shangmian Sun, Shaoqing Wu, Shengfeng Ye, T.~Wang,
  W.~L. Xiao, Wei An, Xianzu Wang, Xiaowen Sun, Xiaoxiang Wang, Ying Tang, Yukun Zha, Zekai Zhang, Zhe Ju, Zhen Zhang, and Zihua Qu.
\newblock Deepseek-v3.2: Pushing the frontier of open large language models, 2025.
\newblock URL \url{https://arxiv.org/abs/2512.02556}.

\bibitem[Glass et~al.(2022)Glass, Rossiello, Chowdhury, Naik, Cai, and Gliozzo]{glass2022re2g}
Michael Glass, Gaetano Rossiello, Md~Faisal~Mahbub Chowdhury, Ankita Naik, Pengshan Cai, and Alfio Gliozzo.
\newblock Re2g: Retrieve, rerank, generate.
\newblock In \emph{Proceedings of the 2022 Conference of the North American Chapter of the Association for Computational Linguistics: Human Language Technologies}, pages 2701--2715, 2022.

\bibitem[{Google}(2025)]{google2025gemini3pro}
{Google}.
\newblock Gemini 3 pro model card.
\newblock Technical report, Google DeepMind, 11 2025.
\newblock URL \url{https://storage.googleapis.com/deepmind-media/Model-Cards/Gemini-3-Pro-Model-Card.pdf}.
\newblock Accessed: 2026-01-06.

\bibitem[Hendrycks et~al.(2021)Hendrycks, Burns, Kadavath, Arora, Basart, Tang, Song, and Steinhardt]{hendrycks2021measuringmathematicalproblemsolving}
Dan Hendrycks, Collin Burns, Saurav Kadavath, Akul Arora, Steven Basart, Eric Tang, Dawn Song, and Jacob Steinhardt.
\newblock Measuring mathematical problem solving with the math dataset, 2021.
\newblock URL \url{https://arxiv.org/abs/2103.03874}.

\bibitem[Ho et~al.(2020)Ho, Nguyen, Sugawara, and Aizawa]{ho2020constructingmultihopqadataset}
Xanh Ho, Anh-Khoa~Duong Nguyen, Saku Sugawara, and Akiko Aizawa.
\newblock Constructing a multi-hop qa dataset for comprehensive evaluation of reasoning steps, 2020.
\newblock URL \url{https://arxiv.org/abs/2011.01060}.

\bibitem[Huang et~al.(2025)Huang, Bu, Zhou, Qu, Liu, Yang, Xu, and Zhao]{huang-etal-2025-empirical}
Hui Huang, Xingyuan Bu, Hongli Zhou, Yingqi Qu, Jing Liu, Muyun Yang, Bing Xu, and Tiejun Zhao.
\newblock An empirical study of {LLM}-as-a-judge for {LLM} evaluation: Fine-tuned judge model is not a general substitute for {GPT}-4.
\newblock In Wanxiang Che, Joyce Nabende, Ekaterina Shutova, and Mohammad~Taher Pilehvar, editors, \emph{Findings of the Association for Computational Linguistics: ACL 2025}, pages 5880--5895, Vienna, Austria, July 2025. Association for Computational Linguistics.
\newblock ISBN 979-8-89176-256-5.
\newblock \doi{10.18653/v1/2025.findings-acl.306}.
\newblock URL \url{https://aclanthology.org/2025.findings-acl.306/}.

\bibitem[Jiang et~al.(2023)Jiang, Xu, Gao, Sun, Liu, Dwivedi-Yu, Yang, Callan, and Neubig]{jiang-etal-2023-active}
Zhengbao Jiang, Frank Xu, Luyu Gao, Zhiqing Sun, Qian Liu, Jane Dwivedi-Yu, Yiming Yang, Jamie Callan, and Graham Neubig.
\newblock Active retrieval augmented generation.
\newblock In Houda Bouamor, Juan Pino, and Kalika Bali, editors, \emph{Proceedings of the 2023 Conference on Empirical Methods in Natural Language Processing}, pages 7969--7992, Singapore, December 2023. Association for Computational Linguistics.
\newblock \doi{10.18653/v1/2023.emnlp-main.495}.
\newblock URL \url{https://aclanthology.org/2023.emnlp-main.495/}.

\bibitem[Jin et~al.(2025{\natexlab{a}})Jin, Yoon, Kargupta, Arik, and Han]{jin2025empiricalstudyreinforcementlearning}
Bowen Jin, Jinsung Yoon, Priyanka Kargupta, Sercan~O. Arik, and Jiawei Han.
\newblock An empirical study on reinforcement learning for reasoning-search interleaved llm agents, 2025{\natexlab{a}}.
\newblock URL \url{https://arxiv.org/abs/2505.15117}.

\bibitem[Jin et~al.(2025{\natexlab{b}})Jin, Zeng, Yue, Yoon, Arik, Wang, Zamani, and Han]{jin2025searchr1trainingllmsreason}
Bowen Jin, Hansi Zeng, Zhenrui Yue, Jinsung Yoon, Sercan Arik, Dong Wang, Hamed Zamani, and Jiawei Han.
\newblock Search-r1: Training llms to reason and leverage search engines with reinforcement learning, 2025{\natexlab{b}}.
\newblock URL \url{https://arxiv.org/abs/2503.09516}.

\bibitem[Joshi et~al.(2017)Joshi, Choi, Weld, and Zettlemoyer]{joshi2017triviaqalargescaledistantly}
Mandar Joshi, Eunsol Choi, Daniel~S. Weld, and Luke Zettlemoyer.
\newblock Triviaqa: A large scale distantly supervised challenge dataset for reading comprehension, 2017.
\newblock URL \url{https://arxiv.org/abs/1705.03551}.

\bibitem[Kazemnejad et~al.(2025)Kazemnejad, Aghajohari, Portelance, Sordoni, Reddy, Courville, and Roux]{kazemnejad2025vinepporefiningcreditassignment}
Amirhossein Kazemnejad, Milad Aghajohari, Eva Portelance, Alessandro Sordoni, Siva Reddy, Aaron Courville, and Nicolas~Le Roux.
\newblock Vineppo: Refining credit assignment in rl training of llms, 2025.
\newblock URL \url{https://arxiv.org/abs/2410.01679}.

\bibitem[Kwiatkowski et~al.(2019)Kwiatkowski, Palomaki, Redfield, Collins, Parikh, Alberti, Epstein, Polosukhin, Devlin, Lee, Toutanova, Jones, Kelcey, Chang, Dai, Uszkoreit, Le, and Petrov]{kwiatkowski-etal-2019-natural}
Tom Kwiatkowski, Jennimaria Palomaki, Olivia Redfield, Michael Collins, Ankur Parikh, Chris Alberti, Danielle Epstein, Illia Polosukhin, Jacob Devlin, Kenton Lee, Kristina Toutanova, Llion Jones, Matthew Kelcey, Ming-Wei Chang, Andrew~M. Dai, Jakob Uszkoreit, Quoc Le, and Slav Petrov.
\newblock Natural questions: A benchmark for question answering research.
\newblock \emph{Transactions of the Association for Computational Linguistics}, 7:\penalty0 452--466, 2019.
\newblock \doi{10.1162/tacl_a_00276}.
\newblock URL \url{https://aclanthology.org/Q19-1026/}.

\bibitem[Li et~al.(2025{\natexlab{a}})Li, Dong, Jin, Zhang, Zhou, Zhu, Zhang, and Dou]{li2025searcho1agenticsearchenhancedlarge}
Xiaoxi Li, Guanting Dong, Jiajie Jin, Yuyao Zhang, Yujia Zhou, Yutao Zhu, Peitian Zhang, and Zhicheng Dou.
\newblock Search-o1: Agentic search-enhanced large reasoning models, 2025{\natexlab{a}}.
\newblock URL \url{https://arxiv.org/abs/2501.05366}.

\bibitem[Li et~al.(2025{\natexlab{b}})Li, Jin, Zhou, Zhang, Zhang, Zhu, and Dou]{li2025matching}
Xiaoxi Li, Jiajie Jin, Yujia Zhou, Yuyao Zhang, Peitian Zhang, Yutao Zhu, and Zhicheng Dou.
\newblock From matching to generation: A survey on generative information retrieval.
\newblock \emph{ACM Transactions on Information Systems}, 43\penalty0 (3):\penalty0 1--62, 2025{\natexlab{b}}.

\bibitem[Li et~al.(2025{\natexlab{c}})Li, Luo, Li, Li, Cheng, Wang, Zheng, Wang, Yin, and Qiu]{li2025r3}
Yuan Li, Qi~Luo, Xiaonan Li, Bufan Li, Qinyuan Cheng, Bo~Wang, Yining Zheng, Yuxin Wang, Zhangyue Yin, and Xipeng Qiu.
\newblock R3-rag: Learning step-by-step reasoning and retrieval for llms via reinforcement learning.
\newblock \emph{arXiv preprint arXiv:2505.23794}, 2025{\natexlab{c}}.

\bibitem[Lightman et~al.(2023)Lightman, Kosaraju, Burda, Edwards, Baker, Lee, Leike, Schulman, Sutskever, and Cobbe]{lightman2023letsverifystepstep}
Hunter Lightman, Vineet Kosaraju, Yura Burda, Harri Edwards, Bowen Baker, Teddy Lee, Jan Leike, John Schulman, Ilya Sutskever, and Karl Cobbe.
\newblock Let's verify step by step, 2023.
\newblock URL \url{https://arxiv.org/abs/2305.20050}.

\bibitem[Lin et~al.(2026)Lin, Hu, Wang, Zhou, Xi, Guo, Liu, Wang, Dou, Zhou, Yan, Han, Gui, Zhang, and Huang]{lin2026mmdocr1trainingagentslong}
Jiahang Lin, Kai Hu, Binghai Wang, Yuhao Zhou, Zhiheng Xi, Honglin Guo, Shichun Liu, Junzhe Wang, Shihan Dou, Enyu Zhou, Hang Yan, Zhenhua Han, Tao Gui, Qi~Zhang, and Xuanjing Huang.
\newblock Mm-doc-r1: Training agents for long document visual question answering through multi-turn reinforcement learning, 2026.
\newblock URL \url{https://arxiv.org/abs/2604.13579}.

\bibitem[Liu et~al.(2024)Liu, Wang, Liu, Zeng, Yan, Sun, Liu, and Zhou]{liu2024improvingmultistepreasoningabilities}
Jiacai Liu, Chaojie Wang, Chris~Yuhao Liu, Liang Zeng, Rui Yan, Yiwen Sun, Yang Liu, and Yahui Zhou.
\newblock Improving multi-step reasoning abilities of large language models with direct advantage policy optimization, 2024.
\newblock URL \url{https://arxiv.org/abs/2412.18279}.

\bibitem[Mallen et~al.(2023)Mallen, Asai, Zhong, Das, Khashabi, and Hajishirzi]{mallen2023trustlanguagemodelsinvestigating}
Alex Mallen, Akari Asai, Victor Zhong, Rajarshi Das, Daniel Khashabi, and Hannaneh Hajishirzi.
\newblock When not to trust language models: Investigating effectiveness of parametric and non-parametric memories, 2023.
\newblock URL \url{https://arxiv.org/abs/2212.10511}.

\bibitem[Nakano et~al.(2022)Nakano, Hilton, Balaji, Wu, Ouyang, Kim, Hesse, Jain, Kosaraju, Saunders, Jiang, Cobbe, Eloundou, Krueger, Button, Knight, Chess, and Schulman]{nakano2022webgptbrowserassistedquestionansweringhuman}
Reiichiro Nakano, Jacob Hilton, Suchir Balaji, Jeff Wu, Long Ouyang, Christina Kim, Christopher Hesse, Shantanu Jain, Vineet Kosaraju, William Saunders, Xu~Jiang, Karl Cobbe, Tyna Eloundou, Gretchen Krueger, Kevin Button, Matthew Knight, Benjamin Chess, and John Schulman.
\newblock Webgpt: Browser-assisted question-answering with human feedback, 2022.
\newblock URL \url{https://arxiv.org/abs/2112.09332}.

\bibitem[Press et~al.(2023)Press, Zhang, Min, Schmidt, Smith, and Lewis]{press2023measuring}
Ofir Press, Muru Zhang, Sewon Min, Ludwig Schmidt, Noah~A Smith, and Mike Lewis.
\newblock Measuring and narrowing the compositionality gap in language models.
\newblock In \emph{Findings of the Association for Computational Linguistics: EMNLP 2023}, pages 5687--5711, 2023.

\bibitem[Sawarkar et~al.(2024)Sawarkar, Mangal, and Solanki]{sawarkar2024blended}
Kunal Sawarkar, Abhilasha Mangal, and Shivam~Raj Solanki.
\newblock Blended rag: Improving rag (retriever-augmented generation) accuracy with semantic search and hybrid query-based retrievers.
\newblock In \emph{2024 IEEE 7th international conference on multimedia information processing and retrieval (MIPR)}, pages 155--161. IEEE, 2024.

\bibitem[Schulman et~al.(2017)Schulman, Wolski, Dhariwal, Radford, and Klimov]{schulman2017proximalpolicyoptimizationalgorithms}
John Schulman, Filip Wolski, Prafulla Dhariwal, Alec Radford, and Oleg Klimov.
\newblock Proximal policy optimization algorithms, 2017.
\newblock URL \url{https://arxiv.org/abs/1707.06347}.

\bibitem[Shao et~al.(2024)Shao, Wang, Zhu, Xu, Song, Bi, Zhang, Zhang, Li, Wu, and Guo]{shao2024deepseekmathpushinglimitsmathematical}
Zhihong Shao, Peiyi Wang, Qihao Zhu, Runxin Xu, Junxiao Song, Xiao Bi, Haowei Zhang, Mingchuan Zhang, Y.~K. Li, Y.~Wu, and Daya Guo.
\newblock Deepseekmath: Pushing the limits of mathematical reasoning in open language models, 2024.
\newblock URL \url{https://arxiv.org/abs/2402.03300}.

\bibitem[Sheng et~al.(2024)Sheng, Zhang, Ye, Wu, Zhang, Zhang, Peng, Lin, and Wu]{sheng2024hybridflow}
Guangming Sheng, Chi Zhang, Zilingfeng Ye, Xibin Wu, Wang Zhang, Ru~Zhang, Yanghua Peng, Haibin Lin, and Chuan Wu.
\newblock Hybridflow: A flexible and efficient rlhf framework.
\newblock \emph{arXiv preprint arXiv: 2409.19256}, 2024.

\bibitem[Singh et~al.(2025)Singh, Fry, Perelman, Tart, Ganesh, El-Kishky, McLaughlin, Low, Ostrow, Ananthram, Nathan, Luo, Helyar, Madry, Efremov, Spyra, Baker-Whitcomb, Beutel, Karpenko, Makelov, Neitz, Wei, Barr, Kirchmeyer, Ivanov, Christakis, Gillespie, Tam, Bennett, Wan, Huang, Sandjideh, Yang, Kumar, Saraiva, Vallone, Gheorghe, Garcia, Braunstein, Liu, Schmidt, Mereskin, Mishchenko, Applebaum, Rogerson, Rajan, Wei, Kotha, Srivastava, Agrawal, Vijayvergiya, Tyra, Nair, Nayak, Eggers, Ji, Hoover, Chen, Chen, Barak, Minaiev, Hao, Baker, Lightcap, McKinzie, Wang, Quinn, Fioca, Hsu, Yang, Yu, Zhang, Brenner, Zetino, Raymond, Lugaresi, Paz, Hudson, Whitney, Li, Chen, Cole, Voss, Ding, Shen, Huang, Colby, Hallacy, Koch, Lu, Kaplan, Kim, Minott-Henriques, Frey, Yu, Czarnecki, Reid, Wei, Decareaux, Scheau, Zhang, Forbes, Tang, Goldberg, Roberts, Palmie, Kappler, Levine, Wright, Leo, Lin, Robinson, Grabb, Chen, Lim, Salama, Bhattacharjee, Tsipras, Li, Yu, Strouse, Williams, Hunn, Bayes, Arbus, Akyurek, Le,
  Widmann, Yani, Proehl, Sert, Cheung, Schwartz, Han, Jiang, Mitchell, Sigler, Wallace, Ritter, Kavanaugh, Mays, Nikishin, Li, Such, de~Avila Belbute~Peres, Raso, Bekerman, Tsimpourlas, Chantzis, Song, Zhang, Raila, McGrath, Briggs, Yang, Parascandolo, Chabot, Kim, Zhao, Valiant, Leclerc, Salman, Wang, Sheng, Jiang, Wang, Jin, Sikchi, Schmidt, Aspegren, Chen, Qiu, Lightman, Covert, Kivlichan, Silber, Sohl, Hammoud, Clavera, Lan, Akkaya, Kostrikov, Kofman, Etinger, Singal, Hehir, Huh, Pan, Wilczynski, Pachocki, Lee, Quinn, Kiros, Kalra, Samaroo, Wang, Wolfe, Chen, Wang, Harb, Han, Wang, Zhao, Chen, Yang, Tworek, Chand, Landon, Liang, Lin, Liu, Wang, Tang, Yin, Jang, Morris, Flynn, Ferstad, Heidecke, Fishbein, Hallman, Grant, Chien, Gordon, Park, Liss, Kraaijeveld, Guay, Mo, Lawson, McGrath, Vendrow, Jiao, Lee, Steele, Wang, Mao, Chen, Hayashi, Xiao, Salahi, Wu, Sekhri, Sharma, Singhal, Li, Nguyen, Gu-Lemberg, King, Liu, Stone, Yu, Ying, Georgiev, Lim, Tirumala, Miller, Ahmad, Lv, Clare, Fauconnet, Itow, Yang,
  Romaniuk, Anise, Byron, Pathak, Maksin, Lo, Ho, Jing, Wu, Xiong, Mamitsuka, Yang, McCallum, Held, Bourgeois, Engstrom, Kuhn, Feuvrier, Zhang, Switzer, Kondraciuk, Kaiser, Joglekar, Singh, Shah, Stratta, Williams, Chen, Sun, Cayton, Li, Zhang, Aljubeh, Nichols, Haines, Schwarzer, Gupta, Shah, Huang, Dong, Wang, Glaese, Carroll, Lampe, Malek, Sharman, Zhang, Wang, Pokrass, Florian, Pavlov, Wang, Chen, Wang, Feng, Bavarian, Lin, Abdool, Rohaninejad, Soto, Staudacher, LaFontaine, Marwell, Liu, Preston, Turley, Ansman, Blades, Pancha, Mikhaylin, Felix, Handa, Rai, Keskar, Brown, Nachum, Boiko, Murk, Watkins, Gleeson, Mishkin, Lesiewicz, Baltescu, Belov, Zhokhov, Pronin, Guo, Thacker, Liu, Yuan, Liu, Dias, Puckett, Arora, Mullapudi, Gaon, Miyara, Song, Aggarwal, Marsan, Yemiru, Xiong, Kshirsagar, Nuttall, Tsiupa, Eldan, Wang, James, Ziv, Shu, Nigmatullin, Jain, Talaie, Altman, Arnesen, Toizer, Toyer, Miserendino, Agarwal, Yoo, Heon, Ethersmith, Grove, Taylor, Bubeck, Banesiu, Amdo, Zhao, Wu, Santurkar, Zhao,
  Chaudhuri, Krishnaswamy, Shuaiqi, Xia, Cheng, Anadkat, Fishman, Tobin, Fu, Jain, Mei, Egoian, Kim, Golden, Mah, Lin, Imm, Sharpe, Yadlowsky, Choudhry, Eum, Sanjeev, Khan, Stramer, Wang, Xin, Gogineni, Christianson, Sanders, Patwardhan, Degry, Shadwell, Fu, Gao, Garipov, Sriskandarajah, Sherbakov, Kaftan, Hiratsuka, Wang, Song, Zhao, Peterson, Kharitonov, Chernova, Kosaraju, Kuo, Pong, Verma, Petrov, Jiang, Zhang, Zhou, Xie, Zhan, McCabe, DePue, Ellsworth, Bain, Thompson, Chen, Qi, Xiang, Shi, Dubois, Yu, Khakbaz, Wu, Qian, Lee, Chen, Zhang, Xiong, Tian, Cha, Bai, Yang, Yuan, Li, Zhang, Yang, Jin, Jiang, Wang, Wang, Liu, Stubenvoll, Dou, Wu, and Wang]{singh2025openaigpt5card}
Aaditya Singh, Adam Fry, Adam Perelman, Adam Tart, Adi Ganesh, Ahmed El-Kishky, Aidan McLaughlin, Aiden Low, AJ~Ostrow, Akhila Ananthram, Akshay Nathan, Alan Luo, Alec Helyar, Aleksander Madry, Aleksandr Efremov, Aleksandra Spyra, Alex Baker-Whitcomb, Alex Beutel, Alex Karpenko, Alex Makelov, Alex Neitz, Alex Wei, Alexandra Barr, Alexandre Kirchmeyer, Alexey Ivanov, Alexi Christakis, Alistair Gillespie, Allison Tam, Ally Bennett, Alvin Wan, Alyssa Huang, Amy~McDonald Sandjideh, Amy Yang, Ananya Kumar, Andre Saraiva, Andrea Vallone, Andrei Gheorghe, Andres~Garcia Garcia, Andrew Braunstein, Andrew Liu, Andrew Schmidt, Andrey Mereskin, Andrey Mishchenko, Andy Applebaum, Andy Rogerson, Ann Rajan, Annie Wei, Anoop Kotha, Anubha Srivastava, Anushree Agrawal, Arun Vijayvergiya, Ashley Tyra, Ashvin Nair, Avi Nayak, Ben Eggers, Bessie Ji, Beth Hoover, Bill Chen, Blair Chen, Boaz Barak, Borys Minaiev, Botao Hao, Bowen Baker, Brad Lightcap, Brandon McKinzie, Brandon Wang, Brendan Quinn, Brian Fioca, Brian Hsu, Brian
  Yang, Brian Yu, Brian Zhang, Brittany Brenner, Callie~Riggins Zetino, Cameron Raymond, Camillo Lugaresi, Carolina Paz, Cary Hudson, Cedric Whitney, Chak Li, Charles Chen, Charlotte Cole, Chelsea Voss, Chen Ding, Chen Shen, Chengdu Huang, Chris Colby, Chris Hallacy, Chris Koch, Chris Lu, Christina Kaplan, Christina Kim, CJ~Minott-Henriques, Cliff Frey, Cody Yu, Coley Czarnecki, Colin Reid, Colin Wei, Cory Decareaux, Cristina Scheau, Cyril Zhang, Cyrus Forbes, Da~Tang, Dakota Goldberg, Dan Roberts, Dana Palmie, Daniel Kappler, Daniel Levine, Daniel Wright, Dave Leo, David Lin, David Robinson, Declan Grabb, Derek Chen, Derek Lim, Derek Salama, Dibya Bhattacharjee, Dimitris Tsipras, Dinghua Li, Dingli Yu, DJ~Strouse, Drew Williams, Dylan Hunn, Ed~Bayes, Edwin Arbus, Ekin Akyurek, Elaine~Ya Le, Elana Widmann, Eli Yani, Elizabeth Proehl, Enis Sert, Enoch Cheung, Eri Schwartz, Eric Han, Eric Jiang, Eric Mitchell, Eric Sigler, Eric Wallace, Erik Ritter, Erin Kavanaugh, Evan Mays, Evgenii Nikishin, Fangyuan Li,
  Felipe~Petroski Such, Filipe de~Avila Belbute~Peres, Filippo Raso, Florent Bekerman, Foivos Tsimpourlas, Fotis Chantzis, Francis Song, Francis Zhang, Gaby Raila, Garrett McGrath, Gary Briggs, Gary Yang, Giambattista Parascandolo, Gildas Chabot, Grace Kim, Grace Zhao, Gregory Valiant, Guillaume Leclerc, Hadi Salman, Hanson Wang, Hao Sheng, Haoming Jiang, Haoyu Wang, Haozhun Jin, Harshit Sikchi, Heather Schmidt, Henry Aspegren, Honglin Chen, Huida Qiu, Hunter Lightman, Ian Covert, Ian Kivlichan, Ian Silber, Ian Sohl, Ibrahim Hammoud, Ignasi Clavera, Ikai Lan, Ilge Akkaya, Ilya Kostrikov, Irina Kofman, Isak Etinger, Ishaan Singal, Jackie Hehir, Jacob Huh, Jacqueline Pan, Jake Wilczynski, Jakub Pachocki, James Lee, James Quinn, Jamie Kiros, Janvi Kalra, Jasmyn Samaroo, Jason Wang, Jason Wolfe, Jay Chen, Jay Wang, Jean Harb, Jeffrey Han, Jeffrey Wang, Jennifer Zhao, Jeremy Chen, Jerene Yang, Jerry Tworek, Jesse Chand, Jessica Landon, Jessica Liang, Ji~Lin, Jiancheng Liu, Jianfeng Wang, Jie Tang, Jihan Yin,
  Joanne Jang, Joel Morris, Joey Flynn, Johannes Ferstad, Johannes Heidecke, John Fishbein, John Hallman, Jonah Grant, Jonathan Chien, Jonathan Gordon, Jongsoo Park, Jordan Liss, Jos Kraaijeveld, Joseph Guay, Joseph Mo, Josh Lawson, Josh McGrath, Joshua Vendrow, Joy Jiao, Julian Lee, Julie Steele, Julie Wang, Junhua Mao, Kai Chen, Kai Hayashi, Kai Xiao, Kamyar Salahi, Kan Wu, Karan Sekhri, Karan Sharma, Karan Singhal, Karen Li, Kenny Nguyen, Keren Gu-Lemberg, Kevin King, Kevin Liu, Kevin Stone, Kevin Yu, Kristen Ying, Kristian Georgiev, Kristie Lim, Kushal Tirumala, Kyle Miller, Lama Ahmad, Larry Lv, Laura Clare, Laurance Fauconnet, Lauren Itow, Lauren Yang, Laurentia Romaniuk, Leah Anise, Lee Byron, Leher Pathak, Leon Maksin, Leyan Lo, Leyton Ho, Li~Jing, Liang Wu, Liang Xiong, Lien Mamitsuka, Lin Yang, Lindsay McCallum, Lindsey Held, Liz Bourgeois, Logan Engstrom, Lorenz Kuhn, Louis Feuvrier, Lu~Zhang, Lucas Switzer, Lukas Kondraciuk, Lukasz Kaiser, Manas Joglekar, Mandeep Singh, Mandip Shah, Manuka
  Stratta, Marcus Williams, Mark Chen, Mark Sun, Marselus Cayton, Martin Li, Marvin Zhang, Marwan Aljubeh, Matt Nichols, Matthew Haines, Max Schwarzer, Mayank Gupta, Meghan Shah, Melody Huang, Meng Dong, Mengqing Wang, Mia Glaese, Micah Carroll, Michael Lampe, Michael Malek, Michael Sharman, Michael Zhang, Michele Wang, Michelle Pokrass, Mihai Florian, Mikhail Pavlov, Miles Wang, Ming Chen, Mingxuan Wang, Minnia Feng, Mo~Bavarian, Molly Lin, Moose Abdool, Mostafa Rohaninejad, Nacho Soto, Natalie Staudacher, Natan LaFontaine, Nathan Marwell, Nelson Liu, Nick Preston, Nick Turley, Nicklas Ansman, Nicole Blades, Nikil Pancha, Nikita Mikhaylin, Niko Felix, Nikunj Handa, Nishant Rai, Nitish Keskar, Noam Brown, Ofir Nachum, Oleg Boiko, Oleg Murk, Olivia Watkins, Oona Gleeson, Pamela Mishkin, Patryk Lesiewicz, Paul Baltescu, Pavel Belov, Peter Zhokhov, Philip Pronin, Phillip Guo, Phoebe Thacker, Qi~Liu, Qiming Yuan, Qinghua Liu, Rachel Dias, Rachel Puckett, Rahul Arora, Ravi~Teja Mullapudi, Raz Gaon, Reah Miyara,
  Rennie Song, Rishabh Aggarwal, RJ~Marsan, Robel Yemiru, Robert Xiong, Rohan Kshirsagar, Rohan Nuttall, Roman Tsiupa, Ronen Eldan, Rose Wang, Roshan James, Roy Ziv, Rui Shu, Ruslan Nigmatullin, Saachi Jain, Saam Talaie, Sam Altman, Sam Arnesen, Sam Toizer, Sam Toyer, Samuel Miserendino, Sandhini Agarwal, Sarah Yoo, Savannah Heon, Scott Ethersmith, Sean Grove, Sean Taylor, Sebastien Bubeck, Sever Banesiu, Shaokyi Amdo, Shengjia Zhao, Sherwin Wu, Shibani Santurkar, Shiyu Zhao, Shraman~Ray Chaudhuri, Shreyas Krishnaswamy, Shuaiqi, Xia, Shuyang Cheng, Shyamal Anadkat, Simón~Posada Fishman, Simon Tobin, Siyuan Fu, Somay Jain, Song Mei, Sonya Egoian, Spencer Kim, Spug Golden, SQ~Mah, Steph Lin, Stephen Imm, Steve Sharpe, Steve Yadlowsky, Sulman Choudhry, Sungwon Eum, Suvansh Sanjeev, Tabarak Khan, Tal Stramer, Tao Wang, Tao Xin, Tarun Gogineni, Taya Christianson, Ted Sanders, Tejal Patwardhan, Thomas Degry, Thomas Shadwell, Tianfu Fu, Tianshi Gao, Timur Garipov, Tina Sriskandarajah, Toki Sherbakov, Tomer Kaftan,
  Tomo Hiratsuka, Tongzhou Wang, Tony Song, Tony Zhao, Troy Peterson, Val Kharitonov, Victoria Chernova, Vineet Kosaraju, Vishal Kuo, Vitchyr Pong, Vivek Verma, Vlad Petrov, Wanning Jiang, Weixing Zhang, Wenda Zhou, Wenlei Xie, Wenting Zhan, Wes McCabe, Will DePue, Will Ellsworth, Wulfie Bain, Wyatt Thompson, Xiangning Chen, Xiangyu Qi, Xin Xiang, Xinwei Shi, Yann Dubois, Yaodong Yu, Yara Khakbaz, Yifan Wu, Yilei Qian, Yin~Tat Lee, Yinbo Chen, Yizhen Zhang, Yizhong Xiong, Yonglong Tian, Young Cha, Yu~Bai, Yu~Yang, Yuan Yuan, Yuanzhi Li, Yufeng Zhang, Yuguang Yang, Yujia Jin, Yun Jiang, Yunyun Wang, Yushi Wang, Yutian Liu, Zach Stubenvoll, Zehao Dou, Zheng Wu, and Zhigang Wang.
\newblock Openai gpt-5 system card, 2025.
\newblock URL \url{https://arxiv.org/abs/2601.03267}.

\bibitem[Song et~al.(2025)Song, Jiang, Min, Chen, Chen, Zhao, Fang, and Wen]{song2025r1searcherincentivizingsearchcapability}
Huatong Song, Jinhao Jiang, Yingqian Min, Jie Chen, Zhipeng Chen, Wayne~Xin Zhao, Lei Fang, and Ji-Rong Wen.
\newblock R1-searcher: Incentivizing the search capability in llms via reinforcement learning, 2025.
\newblock URL \url{https://arxiv.org/abs/2503.05592}.

\bibitem[Team(2024)]{qwen2.5}
Qwen Team.
\newblock Qwen2.5: A party of foundation models, September 2024.
\newblock URL \url{https://qwen.ai/blog?id=qwen2.5}.

\bibitem[Team(2025{\natexlab{a}})]{qwen3max}
Qwen Team.
\newblock Qwen3-max: Just scale it, September 2025{\natexlab{a}}.
\newblock URL \url{https://qwen.ai/blog?id=241398b9cd6353de490b0f82806c7848c5d2777d}.

\bibitem[Team(2025{\natexlab{b}})]{qwen3technicalreport}
Qwen Team.
\newblock Qwen3 technical report, 2025{\natexlab{b}}.
\newblock URL \url{https://arxiv.org/abs/2505.09388}.

\bibitem[Trivedi et~al.(2022)Trivedi, Balasubramanian, Khot, and Sabharwal]{trivedi2022musiquemultihopquestionssinglehop}
Harsh Trivedi, Niranjan Balasubramanian, Tushar Khot, and Ashish Sabharwal.
\newblock Musique: Multihop questions via single-hop question composition, 2022.
\newblock URL \url{https://arxiv.org/abs/2108.00573}.

\bibitem[Trivedi et~al.(2023)Trivedi, Balasubramanian, Khot, and Sabharwal]{trivedi-etal-2023-interleaving}
Harsh Trivedi, Niranjan Balasubramanian, Tushar Khot, and Ashish Sabharwal.
\newblock Interleaving retrieval with chain-of-thought reasoning for knowledge-intensive multi-step questions.
\newblock In Anna Rogers, Jordan Boyd-Graber, and Naoaki Okazaki, editors, \emph{Proceedings of the 61st Annual Meeting of the Association for Computational Linguistics (Volume 1: Long Papers)}, pages 10014--10037, Toronto, Canada, July 2023. Association for Computational Linguistics.
\newblock \doi{10.18653/v1/2023.acl-long.557}.
\newblock URL \url{https://aclanthology.org/2023.acl-long.557/}.

\bibitem[Uesato et~al.(2022)Uesato, Kushman, Kumar, Song, Siegel, Wang, Creswell, Irving, and Higgins]{uesato2022solving}
Jonathan Uesato, Nate Kushman, Ramana Kumar, Francis Song, Noah Siegel, Lisa Wang, Antonia Creswell, Geoffrey Irving, and Irina Higgins.
\newblock Solving math word problems with process-and outcome-based feedback.
\newblock \emph{arXiv preprint arXiv:2211.14275}, 2022.

\bibitem[Wang et~al.(2025{\natexlab{a}})Wang, Qian, Zhong, Chen, Qiu, Huang, Jin, Wang, Wong, and Ji]{wang2025actingreasoningmoreteaching}
Hongru Wang, Cheng Qian, Wanjun Zhong, Xiusi Chen, Jiahao Qiu, Shijue Huang, Bowen Jin, Mengdi Wang, Kam-Fai Wong, and Heng Ji.
\newblock Acting less is reasoning more! teaching model to act efficiently, 2025{\natexlab{a}}.
\newblock URL \url{https://arxiv.org/abs/2504.14870}.

\bibitem[Wang et~al.(2022)Wang, Yang, Huang, Jiao, Yang, Jiang, Majumder, and Wei]{wang2022text}
Liang Wang, Nan Yang, Xiaolong Huang, Binxing Jiao, Linjun Yang, Daxin Jiang, Rangan Majumder, and Furu Wei.
\newblock Text embeddings by weakly-supervised contrastive pre-training.
\newblock \emph{arXiv preprint arXiv:2212.03533}, 2022.

\bibitem[Wang et~al.(2025{\natexlab{b}})Wang, Zheng, An, Ouyang, Cai, Wang, and Wu]{wang2025stepsearchignitingllmssearch}
Ziliang Wang, Xuhui Zheng, Kang An, Cijun Ouyang, Jialu Cai, Yuhang Wang, and Yichao Wu.
\newblock Stepsearch: Igniting llms search ability via step-wise proximal policy optimization, 2025{\natexlab{b}}.
\newblock URL \url{https://arxiv.org/abs/2505.15107}.

\bibitem[Wei et~al.(2025)Wei, Zeng, Li, Brown, Frunza, Deng, Schneider, Nevmyvaka, Zhao, Garcia, and Hong]{wei2025reinforcingmultiturnreasoningllm}
Quan Wei, Siliang Zeng, Chenliang Li, William Brown, Oana Frunza, Wei Deng, Anderson Schneider, Yuriy Nevmyvaka, Yang~Katie Zhao, Alfredo Garcia, and Mingyi Hong.
\newblock Reinforcing multi-turn reasoning in llm agents via turn-level reward design, 2025.
\newblock URL \url{https://arxiv.org/abs/2505.11821}.

\bibitem[Xi et~al.(2025)Xi, Huang, Liao, Huang, Guo, Liu, Zheng, Ye, Zhang, Chen, He, Ding, Li, Chen, Du, Yao, Xu, Chen, Gui, Wu, Zhang, Huang, and Jiang]{xi2025agentgymrltrainingllmagents}
Zhiheng Xi, Jixuan Huang, Chenyang Liao, Baodai Huang, Honglin Guo, Jiaqi Liu, Rui Zheng, Junjie Ye, Jiazheng Zhang, Wenxiang Chen, Wei He, Yiwen Ding, Guanyu Li, Zehui Chen, Zhengyin Du, Xuesong Yao, Yufei Xu, Jiecao Chen, Tao Gui, Zuxuan Wu, Qi~Zhang, Xuanjing Huang, and Yu-Gang Jiang.
\newblock Agentgym-rl: Training llm agents for long-horizon decision making through multi-turn reinforcement learning, 2025.
\newblock URL \url{https://arxiv.org/abs/2509.08755}.

\bibitem[Yang et~al.(2018)Yang, Qi, Zhang, Bengio, Cohen, Salakhutdinov, and Manning]{yang2018hotpotqadatasetdiverseexplainable}
Zhilin Yang, Peng Qi, Saizheng Zhang, Yoshua Bengio, William~W. Cohen, Ruslan Salakhutdinov, and Christopher~D. Manning.
\newblock Hotpotqa: A dataset for diverse, explainable multi-hop question answering, 2018.
\newblock URL \url{https://arxiv.org/abs/1809.09600}.

\bibitem[Yao et~al.(2022)Yao, Chen, Yang, and Narasimhan]{yao2022webshop}
Shunyu Yao, Howard Chen, John Yang, and Karthik Narasimhan.
\newblock Webshop: Towards scalable real-world web interaction with grounded language agents.
\newblock In S.~Koyejo, S.~Mohamed, A.~Agarwal, D.~Belgrave, K.~Cho, and A.~Oh, editors, \emph{Advances in Neural Information Processing Systems}, volume~35, pages 20744--20757. Curran Associates, Inc., 2022.
\newblock URL \url{https://proceedings.neurips.cc/paper_files/paper/2022/file/82ad13ec01f9fe44c01cb91814fd7b8c-Paper-Conference.pdf}.

\bibitem[Zhang et~al.(2026)Zhang, Tan, Huang, Shen, Ma, Ju, Zhang, Wang, Jing, Deng, Sha, Hu, Tong, Jiang, Geng, Ying, Zhang, Yin, Xi, Dou, Gui, Zhang, and Huang]{zhang2026opennoveltyllmpoweredagenticverifiable}
Ming Zhang, Kexin Tan, Yueyuan Huang, Yujiong Shen, Chunchun Ma, Li~Ju, Xinran Zhang, Yuhui Wang, Wenqing Jing, Jingyi Deng, Huayu Sha, Binze Hu, Jingqi Tong, Changhao Jiang, Yage Geng, Yuankai Ying, Yue Zhang, Zhangyue Yin, Zhiheng Xi, Shihan Dou, Tao Gui, Qi~Zhang, and Xuanjing Huang.
\newblock Opennovelty: An llm-powered agentic system for verifiable scholarly novelty assessment, 2026.
\newblock URL \url{https://arxiv.org/abs/2601.01576}.

\bibitem[Zhang et~al.(2025{\natexlab{a}})Zhang, Li, Zhang, Jia, Wang, Guo, Liu, and Zhao]{zhang2025deep}
Wenlin Zhang, Xiaopeng Li, Yingyi Zhang, Pengyue Jia, Yichao Wang, Huifeng Guo, Yong Liu, and Xiangyu Zhao.
\newblock Deep research: A survey of autonomous research agents.
\newblock \emph{arXiv preprint arXiv:2508.12752}, 2025{\natexlab{a}}.

\bibitem[Zhang et~al.(2025{\natexlab{b}})Zhang, Huang, Song, Zhu, Zhang, Zhao, and Zhao]{zhang2025criticsearchfinegrainedcreditassignment}
Yaocheng Zhang, Haohuan Huang, Zijun Song, Yuanheng Zhu, Qichao Zhang, Zijie Zhao, and Dongbin Zhao.
\newblock Criticsearch: Fine-grained credit assignment for search agents via a retrospective critic, 2025{\natexlab{b}}.
\newblock URL \url{https://arxiv.org/abs/2511.12159}.

\bibitem[Zhao et~al.(2026)Zhao, Zhang, Yu, Wang, Geng, Fu, Yang, Zhang, Jiang, and Cui]{zhao2026retrieval}
Penghao Zhao, Hailin Zhang, Qinhan Yu, Zhengren Wang, Yunteng Geng, Fangcheng Fu, Ling Yang, Wentao Zhang, Jie Jiang, and Bin Cui.
\newblock Retrieval-augmented generation for ai-generated content: A survey.
\newblock \emph{Data Science and Engineering}, pages 1--29, 2026.

\bibitem[Zheng et~al.(2024)Zheng, Yin, Xie, Sun, Huang, Yu, Cao, Kozyrakis, Stoica, Gonzalez, Barrett, and Sheng]{zheng2024sglang}
Lianmin Zheng, Liangsheng Yin, Zhiqiang Xie, Chuyue Sun, Jeff Huang, Cody~Hao Yu, Shiyi Cao, Christos Kozyrakis, Ion Stoica, Joseph~E. Gonzalez, Clark Barrett, and Ying Sheng.
\newblock Sglang: Efficient execution of structured language model programs.
\newblock In A.~Globerson, L.~Mackey, D.~Belgrave, A.~Fan, U.~Paquet, J.~Tomczak, and C.~Zhang, editors, \emph{Advances in Neural Information Processing Systems}, volume~37, pages 62557--62583. Curran Associates, Inc., 2024.
\newblock \doi{10.52202/079017-2000}.
\newblock URL \url{https://proceedings.neurips.cc/paper_files/paper/2024/file/724be4472168f31ba1c9ac630f15dec8-Paper-Conference.pdf}.

\bibitem[Zhou et~al.(2024)Zhou, Liu, Li, Jin, Qian, Liu, Li, Dou, Ho, and Yu]{zhou2024trustworthiness}
Yujia Zhou, Yan Liu, Xiaoxi Li, Jiajie Jin, Hongjin Qian, Zheng Liu, Chaozhuo Li, Zhicheng Dou, Tsung-Yi Ho, and Philip~S Yu.
\newblock Trustworthiness in retrieval-augmented generation systems: A survey.
\newblock \emph{arXiv preprint arXiv:2409.10102}, 2024.

\bibitem[Zhu et~al.(2025)Zhu, Wei, Zhao, Wu, Zou, Ran, Wang, Sun, Zhang, and Li]{zhu2025chain}
Dawei Zhu, Xiyu Wei, Guangxiang Zhao, Wenhao Wu, Haosheng Zou, Junfeng Ran, Xun Wang, Lin Sun, Xiangzheng Zhang, and Sujian Li.
\newblock Chain-of-thought matters: improving long-context language models with reasoning path supervision.
\newblock \emph{arXiv preprint arXiv:2502.20790}, 2025.

\end{thebibliography}

\clearpage
\newpage

\appendix
\section*{\centering \LARGE{Appendix}}

\section{Analysis of Failure Trajectories and Credit Assignment Ambiguity}
\label{sec:failure_modes}

In this section, we discuss why round-level credit assignment is unreliable in failed trajectories, complementing the design choice discussed in Section~\ref{sec:weighting_mechanism}.
The key challenge is that failed trajectories typically lack clearly identifiable low-quality rounds that can be reliably attributed to the final failure.
Such ambiguity leads to a high risk of introducing noise when selectively reweighting search rounds. 
Concretely:

\begin{itemize}[leftmargin=*]
    \item \textbf{Redundant/verification queries}: These rounds look redundant but may serve an implicit verification role (re-checking ambiguous evidence). Heuristically marking such rounds as low-quality would risk removing legitimate verification behavior.
    \item \textbf{Cascading Errors from Misinterpreted Evidence}: Later retrieval rounds may be misled by earlier misinterpretations of superficially relevant yet non-supportive documents, and therefore should not be regarded as the primary source of failure.
    \item \textbf{Attempts that failed to obtain key information}: When a search round recalls no useful documents, it usually shows no obvious error — the failure is caused by insufficient corpus coverage or inadequate retriever recall, not a clearly “bad” decision the agent made.
    \item \textbf{Premature overconfidence}: This usually occurs at the answer round, not during intermediate search rounds, and thus falls outside the scope of round-level search credit assignment.
\end{itemize}

A failure pattern that yields a reliably detectable low-quality signal is \textbf{similar-entity confusion}. Because most failure patterns are inherently ambiguous at round granularity, treating failed trajectories differently (uniform contribution)  avoids reinforcing spurious signals and preserves GRPO’s stable outcome-based baseline.

\section{LLM Judge Calibration Details}
\label{cali_details}

To improve transparency, we describe the calibration protocol of our LLM judge and summarize the associated dataset statistics below.

The calibration set contains 97 search rounds sampled from successful GRPO trajectories with diverse trajectory lengths.
Each search round is annotated by a human annotator with respect to two criteria: retrieval utility and reasoning correctness. In addition, the annotator provides a confidence score indicating the certainty of the judgment for each labeled round.

To analyze how judgment quality varies with trajectory length, Table~\ref{tab:cali_satisfy_stats} reports the distribution of search rounds across different trajectory lengths, distinguishing whether each round simultaneously satisfies both retrieval utility and reasoning correctness. This allows us to examine how the quality of search rounds evolves as trajectories become longer.

To further understand the uncertainty structure of the calibration data, Table~\ref{tab:cali_confidence_stats} breaks down the annotation confidence conditioned on whether a search round satisfies both quality criteria.

Overall, among the 89 high-confidence rounds, the LLM judge agrees with the human annotator on 87 rounds; among the remaining 8 medium-confidence cases, agreement occurs in 6 cases, yielding an overall agreement rate of 95.8\%. This demonstrates strong alignment between the LLM judge and human judgments, particularly in high-confidence annotations.

\begin{table}[ht]
\centering
\begin{minipage}[t]{0.48\textwidth}
    \centering
    \begin{tabular}{c|cc}
      \toprule
      \toprule
      \textbf{\makecell{Number of \\ Search Rounds}} 
& \textbf{\makecell{Satisfies Both \\ Criteria}} 
& \textbf{\makecell{Does not \\ Satisfy}} \\
      \midrule
      1 & 5 & 0 \\
      2 & 17 & 3 \\
      3 & 18 & 12 \\
      $\geq$4 & 18 & 24 \\
      \bottomrule
      \bottomrule
    \end{tabular}
    \caption{Quality distribution of search rounds across different trajectory lengths.}
    \label{tab:cali_satisfy_stats}
\end{minipage}
\hfill
\begin{minipage}[t]{0.48\textwidth}
    \centering
    \begin{tabular}{l|ccc}
      \toprule
      \toprule
      \textbf{Confidence} & \textbf{High}
& \textbf{Medium} & \textbf{Low} \\
      \midrule
      Satisfies both criteria & 55 & 3 & 0 \\
      Does not satisfy & 34 & 5 & 0 \\
      \bottomrule
      \bottomrule
    \end{tabular}
    \caption{Confidence distribution across quality labels.}
    \label{tab:cali_confidence_stats}
\end{minipage}
\end{table}

\section{Computational Cost of LLM Judge}
\label{sec:computational_cost}

We provide a brief analysis of the computational cost introduced by the LLM judge during training.
The judge model GPT-oss-120B has 116.8B parameters and adopts a Mixture-of-Experts (MoE) architecture, where only 5.1B parameters are activated per token. According to official deployment guidelines, the model can be served on a single NVIDIA H100 GPU, making it feasible to integrate into the training pipeline.
In practice, incorporating the LLM judge leads to a moderate increase in training cost. Specifically, CW-GRPO training takes 11.7 hours on 5 GPUs (including one GPU dedicated to the judge), compared to 8.8 hours on 4 GPUs for standard GRPO, corresponding to approximately a 33\% increase in wall-clock time.
Overall, while the LLM judge introduces additional computational cost, we find it manageable in our experimental setup.

\section{Prompts}
\label{sec:appendix_prompts}

We present the detailed prompts that guide the behavior of our LLM judge and search agent in this section.

\begin{figure*}[t]
    \centering
    \includegraphics[width=0.98\linewidth]{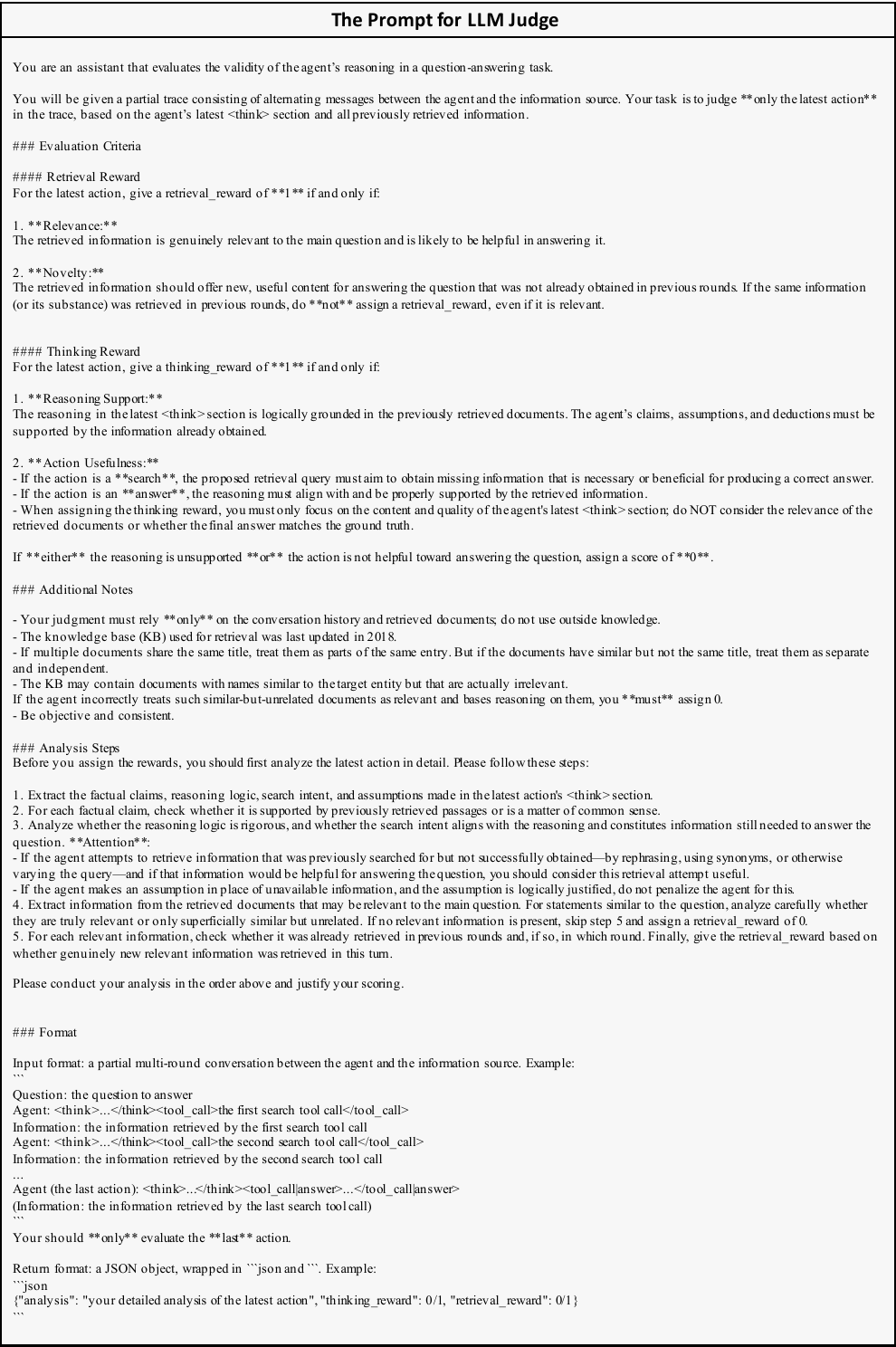}
    \caption{The prompt for LLM Judge to specify the rubric for assessing each search round in terms of retrieval utility and reasoning correctness.}
    \label{fig:llm_judge_prompt}
\end{figure*}

Figure \ref{fig:llm_judge_prompt} illustrates the prompt used by the LLM judge, which specifies the rubrics for assessing each search round in terms of retrieval utility and reasoning correctness, producing the binary contribution signals used in CW-GRPO.

\begin{figure*}[t]
    \centering
    \includegraphics[width=0.98\linewidth]{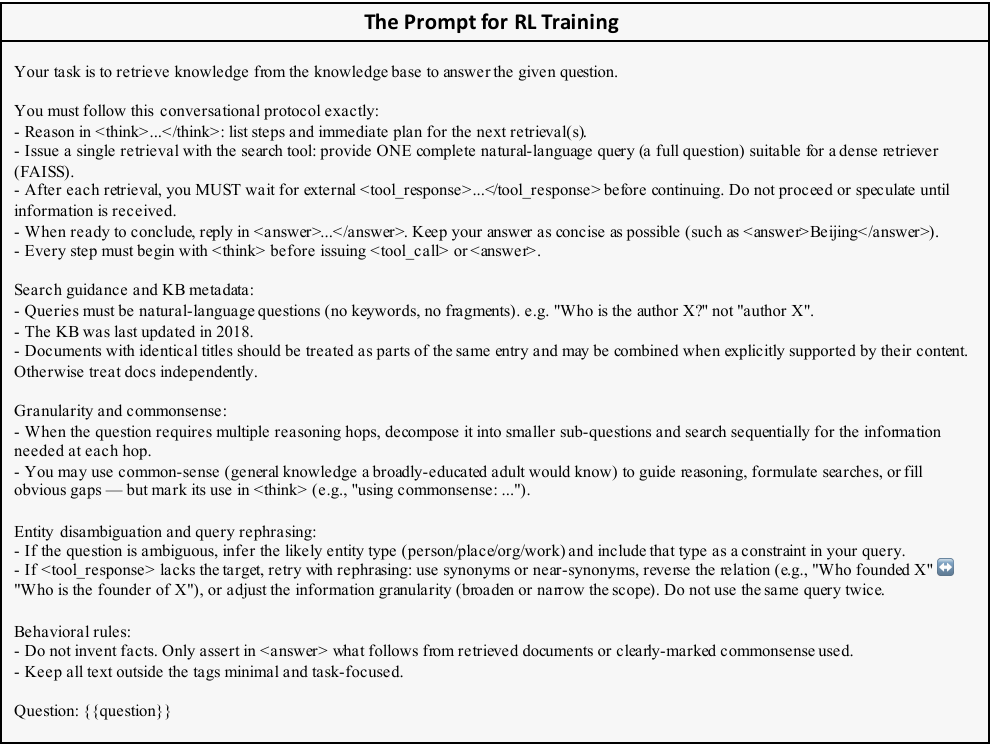}
    \caption{The prompt for RL training.}
    \label{fig:training_prompt}
\end{figure*}

Figure \ref{fig:training_prompt} shows the input prompt used during model training, defining how the search agent structures its reasoning, interacts with the retrieval tool across rounds, and generates the final answer.

\end{document}